\documentclass[lettersize,journal]{IEEEtran}
\usepackage{amsmath,amsfonts}
\usepackage{algorithmic}
\usepackage{algorithm}
\usepackage{array}
\usepackage{textcomp}
\usepackage{stfloats}
\usepackage{url}
\usepackage{verbatim}
\usepackage{graphicx}
\usepackage{cite}
\usepackage{color} 
\usepackage[cmyk]{xcolor}
\usepackage{subfigure}
\usepackage{eso-pic}
\usepackage{multirow}
\usepackage{multicol}
\usepackage{array}
\usepackage{booktabs}
\newcommand{\tabincell}[2]{\begin{tabular}
		{@{}#1@{}}#2\end{tabular}}

\begin{document}

\title{IFAST: Weakly Supervised Interpretable \\Face Anti-spoofing from Single-shot \\Binocular NIR Images}

\author{
Jiancheng Huang$^{*}$,~
Donghao Zhou$^{*}$,~
Shifeng Chen$^{\dagger}$\\
        % <-this % stops a space

\IEEEcompsocitemizethanks{
\IEEEcompsocthanksitem This work is supported by Key-Area Research and Development Program of Guangdong Province (2019B010155003), Shenzhen Science and Technology Innovation Commission (JCYJ20200109114835623, JSGG20220831105002004).
\IEEEcompsocthanksitem Jiancheng Huang and Shifeng Chen are both with ShenZhen Key Lab of Computer Vision and Pattern Recognition, Shenzhen Institute of Advanced Technology, Chinese Academy of Sciences, Shenzhen, 518055, China and also with University of Chinese Academy of Sciences, Beijing, China. (e-mail: jc.huang; shifeng.chen@siat.ac.cn)
\IEEEcompsocthanksitem Donghao Zhou is with Guangdong Provincial Key Laboratory of Computer Vision and Virtual Reality Technology, Shenzhen Institute of Advanced Technology, Chinese Academy of Sciences, Shenzhen, 518055, China, University of Chinese Academy of Sciences, Beijing, China, and also with The Chinese University of Hong Kong, Hong Kong, China. (e-mail: dh.zhou@siat.ac.cn)
}% <-this % stops an unwanted space
\thanks{$^{*}$ Equal contributions.}
\thanks{$^{\dagger}$ Corresponding author.}
}

% \markboth{Journal of \LaTeX\ Class Files,~Vol.~14, No.~8, August~2021}content_tf2(Image.open(original + file).convert('RGB')).unsqueeze(0).to(device)%


\maketitle

\begin{abstract}
\textcolor{black}{Single-shot face anti-spoofing (FAS) is a key technique for securing face recognition systems, and it requires only static images as input. However, single-shot FAS remains a challenging and under-explored problem due to two main reasons: 1) on the data side, learning FAS from RGB images is largely context-dependent, and single-shot images without additional annotations contain limited semantic information. 2) on the model side, existing single-shot FAS models are infeasible to provide proper evidence for their decisions, and FAS methods based on depth estimation require expensive per-pixel annotations. To address these issues, a large binocular NIR image dataset (BNI-FAS) is constructed and published, which contains more than 300,000 real face and plane attack images, and an Interpretable FAS Transformer (IFAST) is proposed that requires only weak supervision to produce interpretable predictions. Our IFAST can produce pixel-wise disparity maps by the proposed disparity estimation Transformer with Dynamic Matching Attention (DMA) block. 
Besides, a well-designed confidence map generator is adopted to cooperate with the proposed dual-teacher distillation module to obtain the final discriminant results. The comprehensive experiments show that our IFAST can achieve state-of-the-art results on BNI-FAS, proving the effectiveness of the single-shot FAS based on binocular NIR images.}
\end{abstract}


\section{Introduction}
\IEEEPARstart{A}{long} \textcolor{black}{with the wide adoption of computer vision and deep learning, there have been significant advances in face recognition in recent years~\cite{li2020review,kortli2020face,wang2023masked}, mainly for identity authentication in many real-world scenarios such as mobile payments, access control, etc. Thus, in addition to the accuracy, the security of face recognition systems also needs to be taken into account. In response to this critical need, \textbf{Face Anti-Spoofing (FAS)} techniques have attracted extensive attention, with the aim of distinguishing between real faces and presentation attacks.}
\begin{figure}[t]
	%\hspace{0.5cm}
	\centering
	\includegraphics[width=0.9\linewidth]{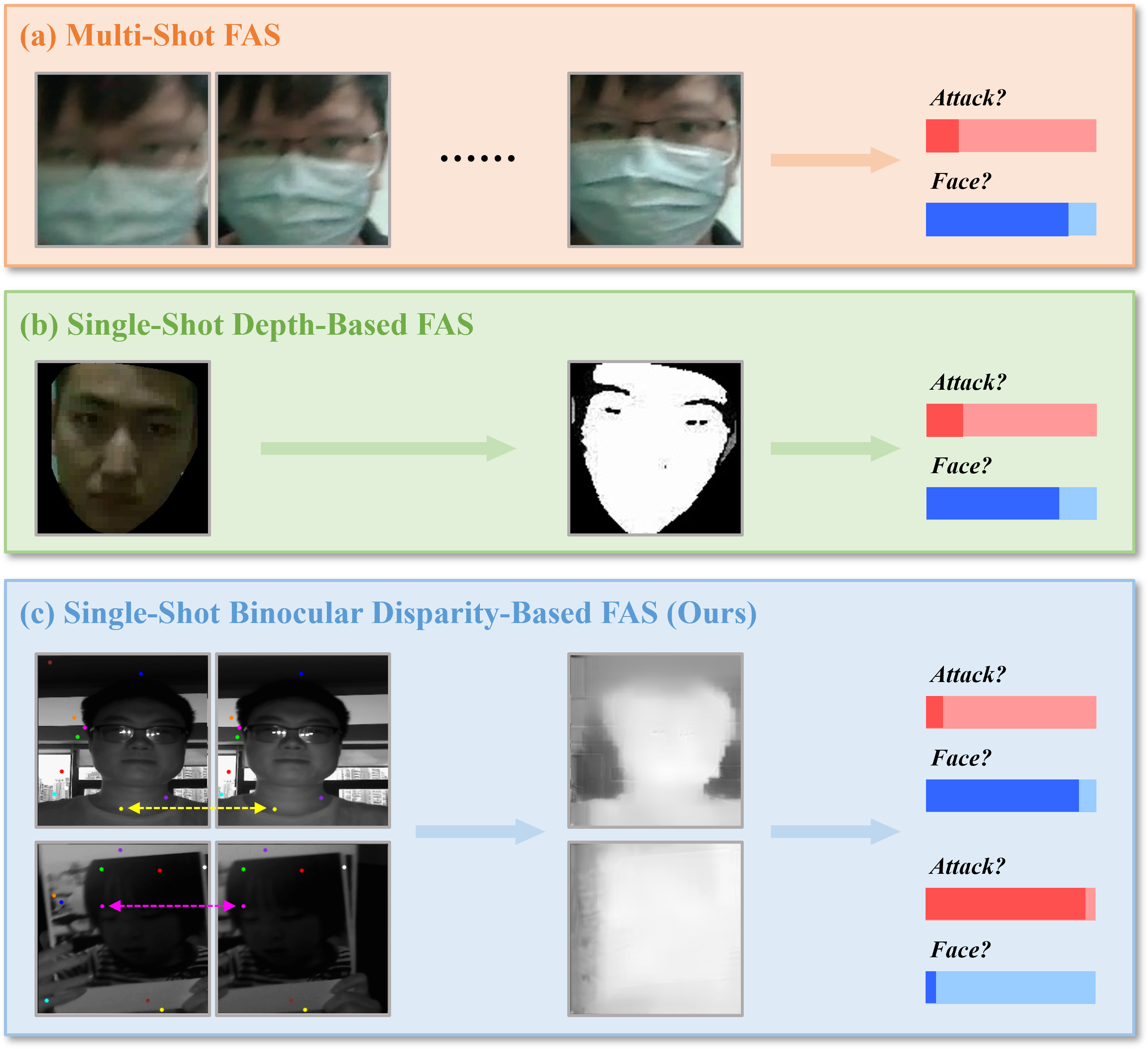}
    % \vspace{-12pt}
	\caption{(a) Multi-shots FAS. (b) Single-shot depth-based FAS. (c) Our single-shot binocular disparity-based FAS. Note that the same colored points are the corresponding maximum attention weight points between the left and right images to emphasize our disparity-based method.}
	\label{TASKfig}
    \vspace{-10pt}
\end{figure}
Most previous FAS methods require a video or a sequence of consecutive images as input~\cite{conf/cvpr/WangYZZQZZL20, TransRPPG,FaceRevelio,DT-Mask}, resulting in time consumption as well as poor user experience. In this paper, we focus on \textbf{single-shot FAS}, where FAS is completed with images obtained through only a single-shot instead of multiple shots (a video or consecutive images), as shown in Fig.~\ref{TASKfig}. \textcolor{black}{We only focus on single-shot FAS because of its obvious benefits. Firstly, single-shot is very fast, and provides a much better user experience compared to shooting a few seconds video~\cite{REHMAN2020103858,wang2022adaptive,wang2022patchnet}. Secondly, single-shot only requires processing of a single captured image. Therefore, its computational cost in the model is much smaller than that of a video-based model~\cite{huang2021multi,LMEG,CNNST,ReplayedVAT,yang2019face,siddiqui,9730902}, which is a necessary requirement when applied to embedded devices.
}
However, single-shot FAS remains a challenging and under-explored problem for the following reasons:
First, in terms of data, 
the most common RGB-image-based methods \cite{REHMAN2020103858,wang2022adaptive,wang2022patchnet,huang2021multi,LMEG} overly rely on the environmental information that is easily affected by brightness, distances, scenes, and attack materials. 
Moreover, due to the lack of binocular and other multi-view FAS datasets, most single-shot methods can only use single-view images without additional annotations, thus only containing limited semantic information and limiting model generalization.
% and thus cause the model to perform FAS by only basic image features, limiting the model generalization.
Therefore, it is of scientific interest to explore other special-modal images in FAS to attain more informative input for models. 
Second, for the model, 
\textcolor{black}{most existing single FAS models have difficulty in providing appropriate evidence for their decisions as they can only output binary results.}
Although there are several works on depth estimation in FAS, these methods either only produce pseudo-depth maps or require expensive per-pixel annotations. 
\textcolor{black}{As we would expect, an ideal FAS model should be able to produce interpretable results with relaxed annotation requirements.}

\begin{figure}[t]
	%\hspace{0.5cm}
	\centering
        \small
	\begin{minipage}[t]{0.25\linewidth}
        \centerline{\includegraphics[width=\linewidth]{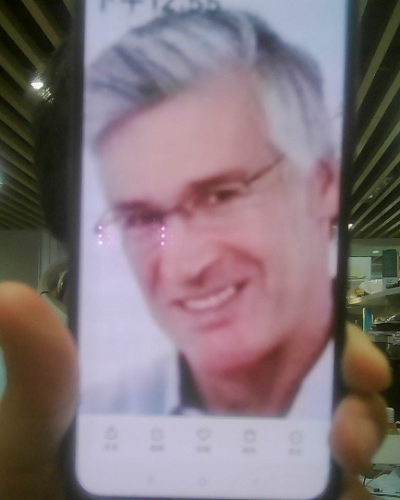}}
        \centerline{(a)}
	\end{minipage}\quad\quad
	\begin{minipage}[t]{0.25\linewidth}
		\centerline{\includegraphics[width=\linewidth]{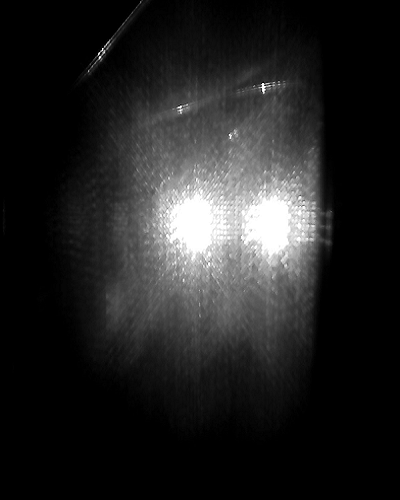}}
        \centerline{(b)}
	\end{minipage}
        % \vspace{-2pt}
	\caption{(a) Photo taken by RGB camera. (b) Photo taken by our NIR camera. The reason why we use NIR cameras is that they are naturally immune to all on-screen attacks, leaving us to consider only paper attacks.} 
	\label{RGBNIR}
    \vspace{-10pt}
\end{figure}

In order to further explore FAS in the data aspect, a \textbf{large-scale binocular near-infrared (NIR) image dataset} is construct and released, which is called \textbf{BNI-FAS}, contains over $300,000$ real-face and plane-attack images for single-shot FAS. 
% To the best of our knowledge, there is currently a lack of publicly available binocular image datasets for FAS, especially those captured by NIR cameras. 
BNI-FAS contains real faces with different illumination and distance, and plane attacks of different materials, which can be used to comprehensively evaluate the performance of FAS methods based on binocular images. \textcolor{black}{BNI-FAS uses only NIR instead of RGB camera for three reasons: 1) NIR reflects all screen attacks, as in Fig. \ref{RGBNIR}. The strong reflection causes all screen attacks to be unrecognizable as faces when captured by the NIR camera, which significantly reduces the cost of anti-spoofing. 2) In many cases, FAS requires senseless shooting at night (i.e. no light is emitted when shooting at night), such as at night or in unlit doorways and vending machines, and NIR cameras are not as sensitive to light as RGB cameras~\cite{yu2023flexible,liu2021face,zhang2019dataset}. Therefore, many existing FAS methods and datasets are developed based on NIR images~\cite{zhang2019dataset,liu2021casia,rostami2021detection}. 3) RGB camera is excluded because IFAST is used to estimate the disparity map by stereo matching between the left and right images. This implies that the binocular images should be of the same modal and therefore NIR should be selected for both sensors of our cameras. Compared to other binocular cameras, structured light cameras, or phone cameras, our industrial binocular NIR cameras are not only lightweight, but also cheap, and which are best suited for embedded devices.}

Moreover, to address the above model issues, \textbf{Interpretable FAS Transformer (IFAST)} is proposed to deal with BNI-FAS. Specifically, inspired by the idea of stereo matching, IFAST generates pixel-wise disparity maps by means of disparity estimation Transformer with dynamic matching attention (DMA) block. Furthermore, a well-designed \textbf{confidence map generator} is used to obtain the final discriminated results. \textcolor{black}{However, the most serious problem is that the labels of the disparity maps cannot be obtained in BNI-FAS because of the extremely high cost of manual labelling. Therefore our method must be trained with only weak supervision. In this paper, weak supervision means training IFAST without labels of disparity maps.}
In order to solve this weak supervision problem, \textbf{dual-teacher distillation module} is proposed to address the lack of pixel-wise manual labels with various losses to achieve the interpretability of results.
Our main contributions are summarized as follows:

\begin{itemize}
	
	\item \textcolor{black}{BNI-FAS, the first and so far the only binocular NIR face images dataset available for single-shot FAS, is constructed and released.} BNI-FAS contains over $300,000$ real-face and plane-attack images in different environments. 
	\item \textcolor{black}{Interpretable FAS Transformer (IFAST) with only weak supervision required is proposed to attain interpretable FAS with limited annotations. This aims to handle the lack of disparity labels for binocular NIR images in real-world scenes.}
	\item Comprehensive experiments show that IFAST can achieve state-of-the-art performance on BNI-FAS when compared to the existing methods, demonstrating the practicality of binocular NIR images in FAS. 
\end{itemize}
% \vspace{-4pt}
\section{Related Work}

\subsection{Face Anti-Spoofing}

\textcolor{black}{Traditional single-shot FAS methods are based on basic image details and handcrafted features for discrimination~\cite{ABTS,DRL,patch-based_CNN,journals/corr/abs-2011-08019,9868051,9796574,9732458,9646915}. 
Image distortion analysis \cite{Wen} uses single-shot input to design statistical features, and color texture analysis \cite{boulkenafet2016face} uses HSV face multi-level LBP feature and YCbCr face LPQ feature. There are also many studies exploring continuous multi-frame images as input which we call multi-shots in Fig.~\ref{TASKfig}~\cite{CNNST,ReplayedVAT,yang2019face,siddiqui,9730902,conf/cvpr/WangYZZQZZL20, TransRPPG,FaceRevelio,DT-Mask}. However, these methods always suffer from high computational cost, which cannot be applied to embedded devices~\cite{yu2022deep,ming2020survey}.
For single-shot, most CNN-based methods utilize end-to-end binary cross-entropy supervision \cite{DeepColorFASD,LGID,patel2016secure,menotti2015deep} with CNN to extract and classify the texture features. However, many FAS methods only output a predicted probability without any additional evidence for their decision \cite{DeepColorFASD,LGID,LucenaJMSVL17,menotti2015deep}.} 

\textcolor{black}{Hence, some methods use auxiliary pixel-wise supervision to enhance interpretability \cite{DCCDN,Liu_2019_CVPR,conf/cvpr/YuZWQ0LZZ20,yu2020fas}. In particular, Atoum et al.~\cite{atoum2017face} are the first to consider depth maps in FAS, as faces on screen or paper are generally flat, and the 3D distribution of real faces differs from that of plane attacks. Some depth-based CNNs such as~\cite{9646915} employ two-stream networks to estimate depth from a single image and extract features from patches. Yu et al.~\cite{conf/cvpr/YuZWQ0LZZ20} propose a special central difference convolution (CDCN) to obtain more discriminative features at the gradient level. Based on~\cite{conf/cvpr/YuZWQ0LZZ20}, DC-CDN~\cite{DCCDN} implements a sparse version of CDCN. In addition to depth, mask, zeromap and RGB input can also act for pixel-wise supervision~\cite{liu2020disentangling,liu2020physics,qin2021meta}. Due to the poor generalization performance of single-shot methods, some works focus on the generalization to unseen domain~\cite{liu2022feature,wang2022domain,jia2021unified}, but they don't completely solve the problem of single-shot FAS due to data limitations.}

\textcolor{black}{Thus, some works start to solve this generalization problem by using special single-shot image such as dual-pixel images. Wu et al.~\cite{wudual} propose a depth estimation method for FAS based on a dual-pixel camera that uses CNN to estimate depth from dual-pixel images. Kang et al.~\cite{kang2022facial} also collect a dual-pixel dataset and use structured light for their depth labeling. However, their methods require expensive manual labeling and are uninterpretable because they estimates depth based on texture context without using stereo matching from a 3D vision perspective.
In summary, the above methods either estimate depth from a single RGB image or rely solely on full-supervised labels, so none of the depths they obtain can be interpreted in terms of stereo geometry.
Without any pixel-wise supervision required, our IFAST performs disparity estimation based on the rule of stereo matching, allowing for better interpretability.}
% \vspace{-5pt}
\begin{figure*}[t]
	\begin{center}
		\includegraphics[width=\linewidth]{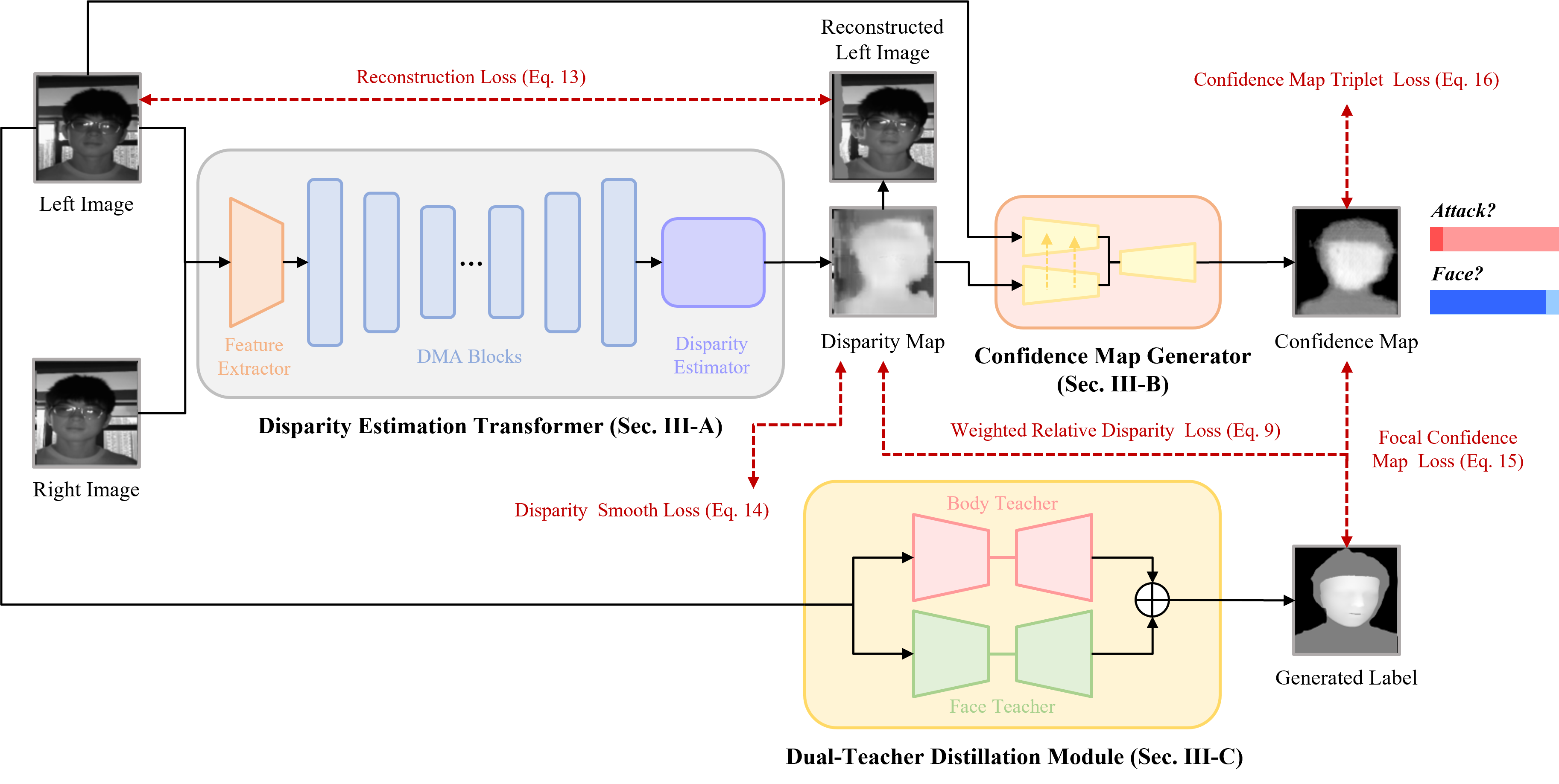}
	\end{center}
    % \vspace{-8pt}
	\caption{The training pipeline of the proposed Interpretable FAS Transformer (IFAST). The input of IFAST is a binocular NIR face image (left and right image). First, an estimated disparity map is obtained by the disparity estimation Transformer. Then, the disparity map and the left image are fed to the confidence map generator. The confidence maps in terms of real face and plane attack are used to complete the classification of FAS. The proposed dual-teacher distillation module is used to support the weakly supervised training.}
	\label{net}
    \vspace{-5pt}
\end{figure*}

\subsection{Depth/Disparity Estimation}\label{re:depth}

Traditionally, this problem can be solved by four modules, i.e., feature extraction, feature matching, disparity calculation, and disparity refinement. 
Small-motion-based methods use vibrating motion to compute depth information~\cite{ha2016high}, where the input images correspond to frames of the input video. For Fast Cost-volume Filtering (CVF)~\cite{hosni2012fast}, a cost volume is constructed and winner-take-all label selection is used to generate a disparity map. Besides, deep learning methods based on stereo matching is used to compute the features of patches through a convolutional neural network to evaluate the similarity between two patches and match them by similarity \cite{ZbontarL15,conf/cvpr/ZagoruykoK15}. Kendall et al.~\cite{kendall2017end} build a 4D feature volume by concatenating features with different disparities and computing the matching cost by 3D convolutions. 
Furthermore, 3D convolution is applied for feature matching and aggregation of matching costs in many similar stereo depth estimation methods~\cite{yang2019hierarchical,yang2020cost,cheng2020hierarchical,gu2020cascade}. 

Notably, STereo TRansformer (STTR) and other attention-based methods replace cost volume construction with dense pixel matching using attention in the Transformer, which relaxes the limitation of the fixed disparity range~\cite{journals/corr/abs-2011-02910, rao2022sliding, bengana2022seeking,Li_2022_CVPR}. \textcolor{black}{Nevertheless, little work has been done to explore depth or disparity estimation of near-infrared (NIR) images. Existing depth estimation methods are not specialized for NIR images and therefore do not work well with NIR images. When applied directly to NIR binocular images, both supervised methods based on synthetic datasets~\cite{journals/corr/abs-2011-02910, rao2022sliding, bengana2022seeking,Li_2022_CVPR} and unsupervised methods based on prior losses~\cite{wang2020parallax} perform poorly, because weak textures, aperture problems and repeated texture problems are more apparent in real NIR images, making it extremely difficult to obtain correct matches. In contrast to them, our IFAST is specifically designed in both model and losses for real binocular NIR images without depth labels. Despite the lack of full supervision, IFAST can still produce interpretable results and achieve state-of-the-art FAS performance on the proposed NIR image datasets.}
% \vspace{-3pt}

\section{Methodology}
\textcolor{black}{As shown in Fig. \ref{net}, The proposed Interpretable FAS Transformer (IFAST) consists of two parts, disparity estimation Transformer and confidence map generator. The former is for obtaining disparity, and the latter is for the classification of FAS. IFAST takes a binocular NIR face image (consisting of a left and a right image). The disparity estimation Transformer estimates a disparity map, which is then fed into the confidence map generator with the left image. The resulting confidence maps for the real face and the plane attack are used to classify. To support weakly supervised training, the dual-teacher distillation module with 5 loss functions is employed.}
We start by introducing the disparity estimation Transformer in Sec.~\ref{sec:3.1}.
Then, we introduce the well-designed confidence map generator and dual-teacher distillation module in Sec.~\ref{sec:3.2} and Sec.~\ref{sec:3.3}, respectively.
Finally, we describe the adopted weak supervised losses in Sec.~\ref{sec:3.4}.

\begin{figure*}[t]
	%\hspace{0.5cm}
	\centering
	\subfigure[]{\includegraphics[width=0.55\linewidth]{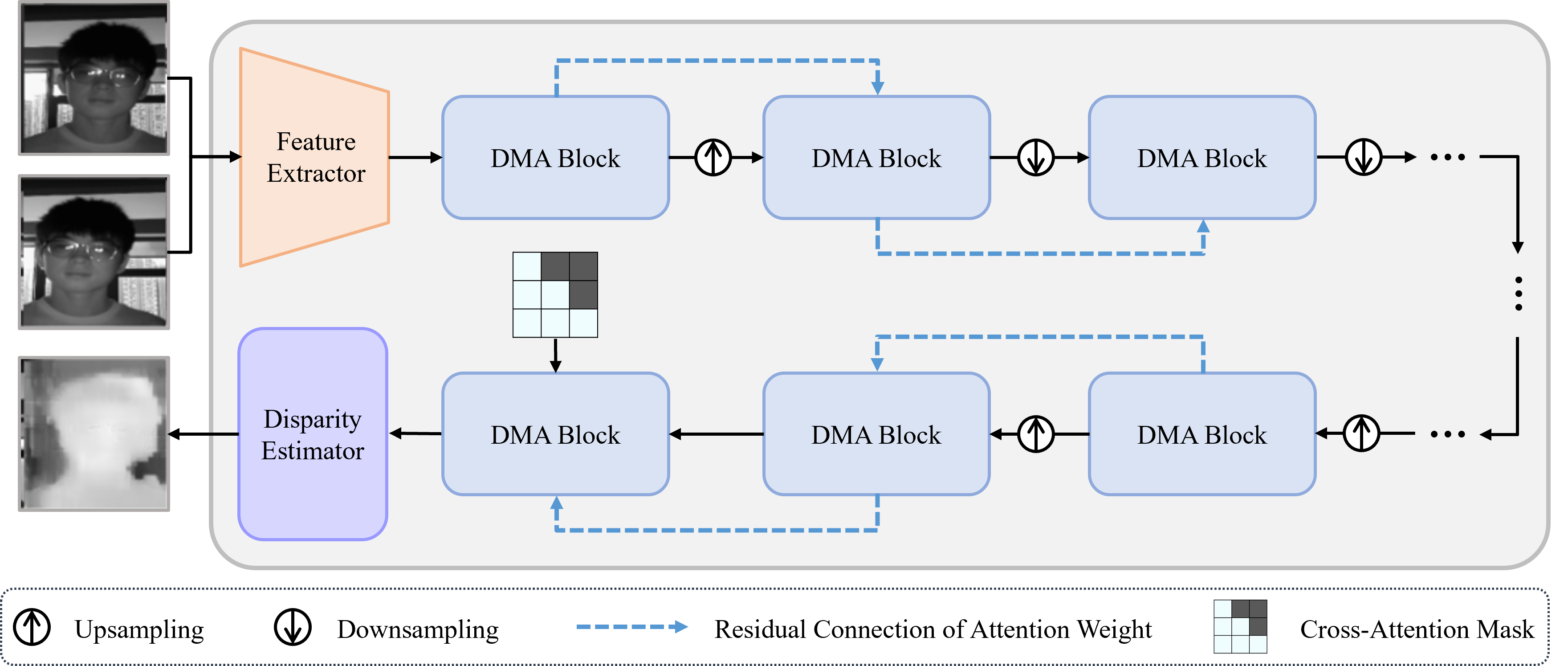}}
	\subfigure[]{\includegraphics[width=0.44\linewidth]{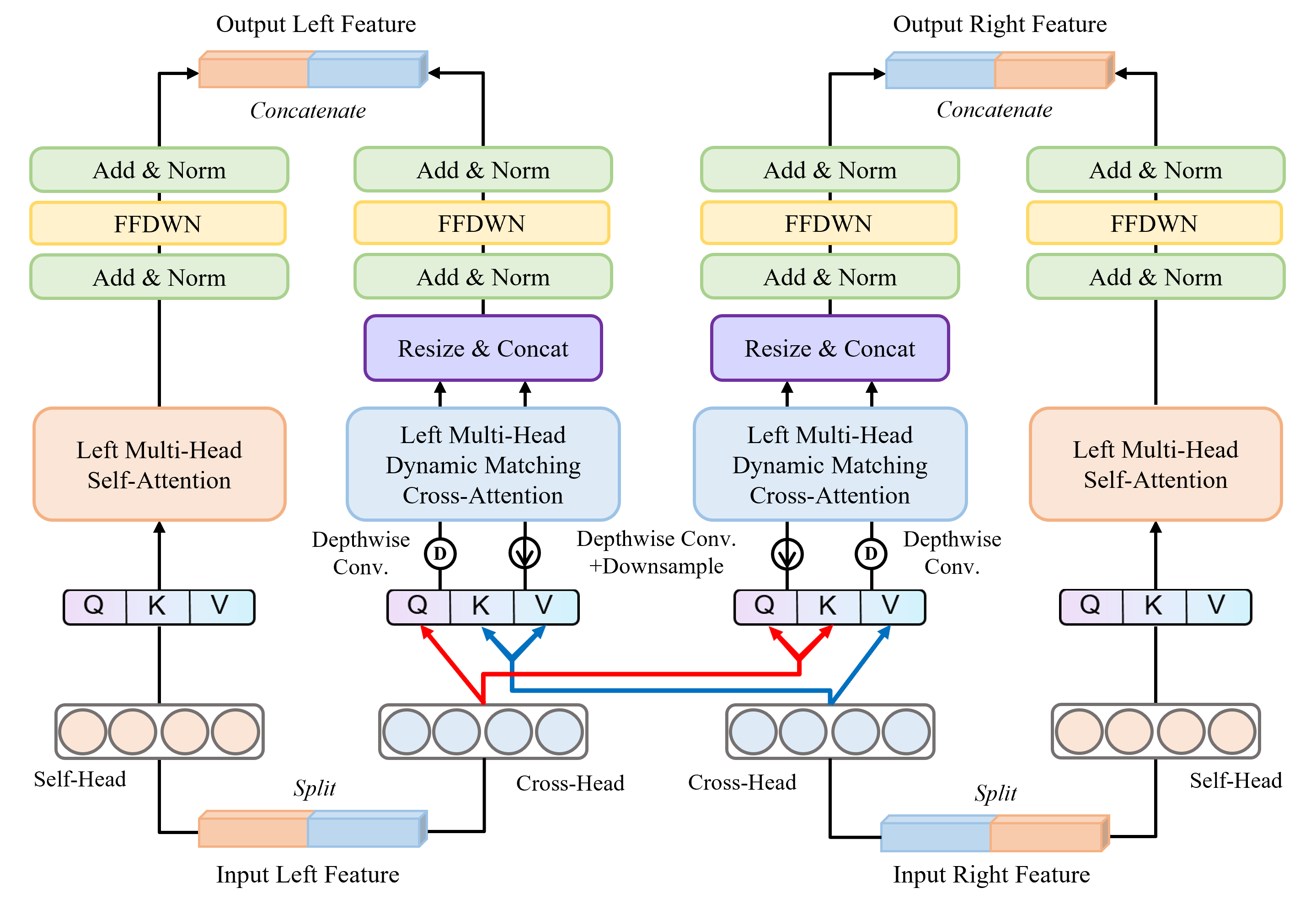}}
    % \vspace{-10pt}
	\caption{(a) The proposed Disparity Estimation Transformer. (b) The Dynamic Matching Attention (DMA) block.} 
	\label{hc}
    % \vspace{-10pt}
\end{figure*}

% \vspace{-5pt}
\subsection{Disparity Estimation Transformer}
\label{sec:3.1}
The architecture of our disparity estimation Transformer follows the normal architecture of Transformer \cite{vaswani2017attention, dosovitskiy2020image, journals/corr/abs-2011-02910}, while it additionally adopts a pivotal technique named dynamic matching attention block. 
In this section, dynamic matching attention block is introduced, followed by a brief introduction to the feature extractor and disparity estimator.

\subsubsection{Feature Extractor} \label{fe}
A feature extractor similar to the hourglass-shaped one in~\cite{liu2020extremely} is used. The basic block contains two convolutional layers with residual connections for efficient global contextualization. $H$ and $W$ denote the height and width of the input images, respectively, and the size of the feature maps obtained by the feature extractor is $C\times \frac{H}{4} \times \frac{W}{4}$. Then, each small patch with a size of $4\times4$ is embedded in a vector with $C$ channels to encode the local and global position-related information.

\subsubsection{Dynamic Matching Attention (DMA) Block}
The attention mechanism \cite{vaswani2017attention} is to compute the attention weights between the sequence elements and is mainly divided into self-attention and cross-attention. Specifically, self-attention is computed along horizontal pixels in the same image, and the relationship between the left and right images is measure with cross-attention. 

Unlike other Transformers \cite{vaswani2017attention, dosovitskiy2020image, journals/corr/abs-2011-02910}, with respect to this DMA block, a particular approach is adopted to compute cross-attention. Transformer~\cite{vaswani2017attention} and STTR\cite{journals/corr/abs-2011-02910} compute self-attention and cross-attention alternately, while PASMNet \cite{wang2020parallax} only computes cross-attention. However, in our method, different heads focus on extracting different features, which can improve the feature representation capability \cite{Ren_2022_CVPR,https://doi.org/10.48550/arxiv.2204.03645,https://doi.org/10.48550/arxiv.2105.08050,Tu_2022_CVPR,https://doi.org/10.48550/arxiv.2106.07631}. Specifically, in a DMA block, half of the heads for self-attention focus on information flow within the same image, while the other half of the heads for cross-attention focus on feature matching between left and right images. This design plays an important role under weak supervision without depth labels, and aims to prevent the chaotic flow between the left and right features in the early stages of training, as detailed in Sec. \ref{sec:abs}.

It is also worth noting that such a split of the heads can also facilitate dynamic matching. \textbf{Dynamic matching} means that the matching window between the left and right images can change dynamically. According to \cite{conf/cvpr/ZagoruykoK15,journals/spl/ParkL17}, the size of the matching window should change dynamically to accommodate the pixel correspondence between the left and right images. However, in STTR~\cite{journals/corr/abs-2011-02910}, the sizes of the patches in all layers are the same, so that the sequence size is constant for all Transformer layers. Moreover, the number of channels $C$ in STTR \cite{journals/corr/abs-2011-02910} is also fixed. 
As shown in Fig.~\ref{hc}(a), dynamic matching is implemented in two ways. First, in the DMA block, Depthwise Convolution (DWConv)~\cite{chollet2017xception} with down-sampling is employed to change the scales in the cross-attention heads. Different scales are accompanied by different matching windows, thus the cross-attention at different scales can be computed in this way. The feature maps are then resized to the original resolution and concatenated. Second, up-sampling and down-sampling among the DMA blocks are added. Down-sampling concatenates the features of four adjacent patches, halving the resolution, and then doubles the feature channels through a linear layer. Up-sampling is the opposite of the above. As a result, the final output feature map has the same shape as the input.

In particular, a cross-attention mask is added to the last DMA block to prevent it from matching wrong positions. In addition, to better transfer attention information and promote gradient back-propagation of attention weights, \textbf{residual connection of attention weight} between adjacent blocks is adopted. Moreover, Feed Forward DWConv Net (FFDWN) is used to extract feature information between channels, unlike~\cite{journals/corr/abs-2011-02910}. 
Besides, the multi-head attention mechanism is also utilized in the DMA block. The feature vector with $C_e$ channels is divided into $N_h$ groups along the channel dimension as $C_h = C_e/N_h$, where $C_h$ is the channel dimension of each attention head and $N_h$ is the number of attention heads. Notably, there are $\frac{N_h}{2}$ heads for cross-attention and $\frac{N_h}{2}$ heads for self-attention. 
To be specific, the overview of our DMA block is depicted in Fig.~\ref{hc}(b).

\begin{figure*}[t]
	\begin{center}
		\includegraphics[width=0.95\linewidth]{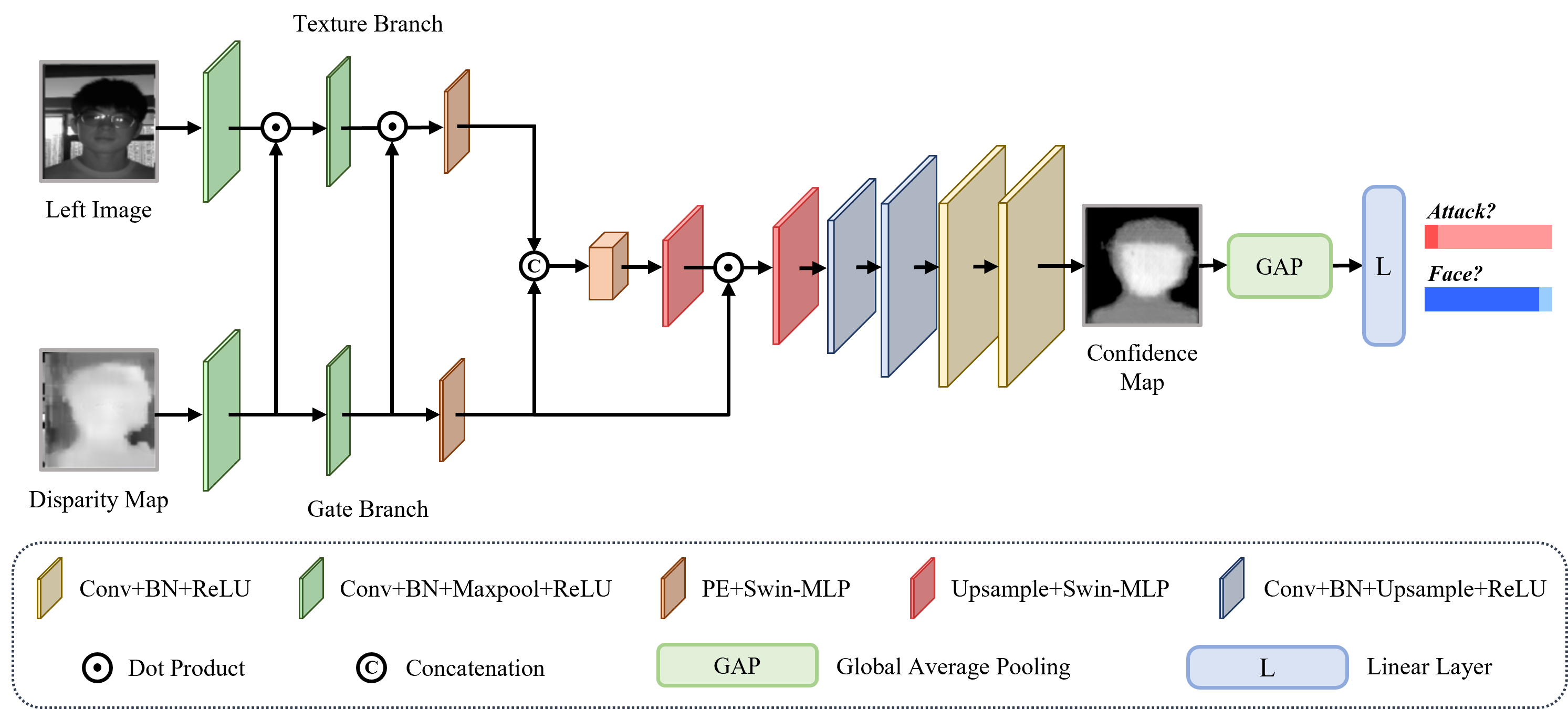}
	\end{center}
        % \vspace{-13pt}
	\caption{The overview of the proposed Confidence Map Generator (CMG).}
	\label{cm}
    % \vspace{-10pt}
\end{figure*}

For each attention head $h$, the query vector $Q_h$, the key vector $K_h$, and the value vector $V_h$ are computed using linear projections. Then, DWConvs are used to obtain different scales of $Q_h$, $K_h$, and $V_h$.  
Taking the left multi-head dynamic matching cross-attention as an example, $E_{L_h}^{\mathrm{cross}}, E_{R_h}^{\mathrm{cross}} \in \mathbb{R}^{BH_h\times W_h \times C_h}$ represent the left and right embeddings in the $h$-th head for computing cross-attention between the left and right images, respectively. Among them, $B$, $H_h$, and $W_h$ denote the batch size, height, and width of the embedding, respectively. In particular, $W_h$ also indicates the sequence length of Transformer. Given the definitions above, the process is shown as follows:
\begin{equation}
\begin{aligned}
&Q_h=W_{Q_h}E_{L_h}^{\mathrm{cross}}+b_{Q_h},\\
&K_h=W_{K_h}E_{R_h}^{\mathrm{cross}}+b_{K_h},\\
&V_h=W_{V_h}E_{R_h}^{\mathrm{cross}}+b_{V_h},\\
\end{aligned}
\end{equation}
where $W_{Q_h}, W_{K_h}, W_{V_h}\in \mathbb{R}^{C_h\times C_h}$ and $b_{Q_h}, b_{K_h}, b_{V_h}\in \mathbb{R}^{C_h}$ are the weights and biases in the $h$-th head. Then DWconvs are used to obtain different scales: 
\begin{equation}
\begin{aligned}
&Q_h^{(1)}=\mathrm{DWConv}^{(1)}(Q_h),\ Q_h^{(2)}=\mathrm{DWConv}^{(2)}(Q_h),\\
&K_h^{(1)}=\mathrm{DWConv}^{(1)}(K_h),\ K_h^{(2)}=\mathrm{DWConv}^{(2)}(K_h),\\
&V_h^{(1)}=\mathrm{DWConv}^{(1)}(V_h),\ V_h^{(2)}=\mathrm{DWConv}^{(2)}(V_h), \\
\end{aligned}
\end{equation}
where $\mathrm{DWConv}^{(1)}$ and $\mathrm{DWConv}^{(2)}$ are the DWConvs with stride 1 and stride 2, respectively. For the sake of brevity, the transformation between the 1D sequence and the 2D spatial structure is omitted. Next, the attention weights of each scale are computed by:
% \vspace{-3pt}
\begin{equation}\label{attn_res}
\begin{aligned}
    \alpha_h^{(1)}=\mathrm{Softmax}(\frac{{Q_h^{(1)}}^TK_h^{(1)}}{\sqrt{C_h}}+\mathrm{Resize}(\alpha_{h-\mathrm{res}}^{(1)})),\\
    \alpha_h^{(2)}=\mathrm{Softmax}(\frac{{Q_h^{(2)}}^TK_h^{(2)}}{\sqrt{C_h}}+\mathrm{Resize}(\alpha_{h-\mathrm{res}}^{(2)})),
\end{aligned}
\end{equation}
where $\alpha_{h}^{(1)} \in \mathbb{R}^{BH_h \times W_h\times W_h}$ and $\alpha_{h}^{(2)} \in \mathbb{R}^{B\frac{H_h}{2} \times \frac{W_h}{2}\times \frac{W_h}{2}}$. $\alpha_{h-\mathrm{res}}^{(1)}$ denotes the residual attention weight from the same position in the previous DMA block. The output value vector $V^O$ can be computed as:
\begin{equation}
\begin{aligned}
&V_h^O = \mathrm{Cat}(\alpha_{h}^{(1)}V_h^{(1)}, \mathrm{Resize}(\alpha_{h}^{(2)}V_h^{(2)})),\\
&V^O=W^O \mathrm{Cat}(V_1^O,...,V_{\frac{N_{h}}{2}}^O)+b^O,
\end{aligned}
\end{equation}
where $W^O \in \mathbb{R}^{C_e \times \frac{C_e}{2}}$ and $b^O \in \mathbb{R}^{\frac{C_e}{2}}$. The output value vector $V^O \in \mathbb{R}^{BH_h\times W_h\times \frac{C_e}{2}}$ is then added to the original feature map for residual connection. Then it is passed through the FFDWN, consisting of two linear layers and a DWconv layer. Finally, the updated feature map is obtained as follows:
\begin{equation}
E_{L}^{\mathrm{cross}}= E_{L}^{\mathrm{cross}}+\mathrm{FFDWN}(E_{L}^{\mathrm{cross}}+V^O),
\end{equation}
\begin{equation}
E_{L}= \mathrm{Cat}(E_{L}^{\mathrm{cross}},E_{L}^{\mathrm{self}}).
\end{equation}

\subsubsection{Disparity Estimator} \label{de}

After the disparity estimation Transformer, the attention weights $a\in \mathbb{R}^{\frac{H}{4}\times\frac{W}{4}\times\frac{W}{4}}$ of the downsampled feature map are obtained by performing softmax processing based on the left image. Subsequently, and these attention weights are used to evaluate the disparity map of the left image. For a given pixel position $(x_l,y_l)$ in the left image, the attention weight vector $a^{x_l,y_l} \in \mathbb{R}^{\frac{W}{4}}$ between this position and all pixels in the same row of the right image is first utilized to find the position with the largest attention weight $(x_r^{\max}, y_l)$. Then, with reference to~\cite{journals/corr/abs-2011-02910}, the positions of its left and right neighbors are taken to form a 3px window. The attention weights of this 3px window are used as the weights of their disparity values, and the disparity $d$ is calculate by weighted summation:
\begin{equation}
x_r^{\max} = \arg\max_{x_r} a^{x_l,y_l}_{x_r},\\  
\end{equation}
\begin{equation}
\begin{aligned}
	&d_{x_l,y_l}= a^{x_l,y_l}_{x^{\max}_r-1}(x_l-(x^{\max}_r-1))+\\&a^{x_l,y_l}_{x^{\max}_r}(x_l-x^{\max}_r)+a^{x_l,y_l}_{x^{\max}_r+1}(x_l-(x^{\max}_r+1)).
\end{aligned}
\end{equation}

Subsequently, the resultant low-scale disparity map is multiplied by the corresponding scale coefficient and then up-sampled to the original image scale. The raw disparity map is normalized and then fed into several ResBlocks~\cite{he2016residual} with the aim of obtaining a smoother disparity map.

% \vspace{-5pt}
\subsection{Confidence Map Generator}
\label{sec:3.2}
Unfortunately, using only the disparity map for FAS results in the loss of all texture information. Inspired by previous work based on pseudo-depth \cite{DCCDN,Liu_2019_CVPR,conf/cvpr/YuZWQ0LZZ20,yu2020fas}, features can be extracted from the disparity map as the gating of the left image to generate the confidence map, and thus design a Confidence Map Generator (CMG) for final classification. 
The size of the confidence map is $2\times H \times W$. The values of these two channels in the confidence map represent the probability that the position belongs to a real face and a plane attack, respectively. As shown in Fig. \ref{cm}, a dual-branch gated network is designed, where the encoder has two branches corresponding to the disparity map and the left image respectively. Each branch consists of 2 convolution blocks and 2 Swin-MLP~\cite{liu2022swin} blocks with down-sample. The outputs of two branches perform a one-to-one dot product for gating. Finally, a more discriminative confidence map can be obtained by several Swin-MLP blocks. Finally, a linear computation is carried out after global average pooling of the confidence map to obtain the logits of the classification.
% \vspace{-8pt}
\subsection{Dual-teacher Distillation Module}
\label{sec:3.3}
Ground-truth depth maps of these images are difficult to obtain, and it is also difficult for human annotators to mark true disparity. This stems from the fact that these binocular images are captured with a binocular NIR camera in real scenes. Nevertheless, the relative depth information is more readily available than ground truth information~\cite{chen2016single}. 
% We believe that it is time-consuming and unnecessary to manually label such a large number of point-to-point data. 
Based on the idea of knowledge distillation, a method for the knowledge distillation of labels is proposed (in Fig.~\ref{net}). First of all, the 3D face reconstruction depth maps are obtained based on the left images by 3DDFA-V2~\cite{conf/eccv/GuoZYYLL20}. 
% It is worth noting that, many other pseudo-depth-based face anti-spoofing methods \cite{conf/cvpr/YuZWQ0LZZ20,conf/cvpr/WangYZZQZZL20,conf/icassp/Peng0ZG20} also adopt 3D face reconstruction to obtain labels. 
However, this kind of label is not enough, as it only has information about a small area of the face but not about any other ones.
% , which means that it is impossible to learn disparity maps with such labels. 
Therefore, the segmentation map of the human body area of each left image is extracted by PP-HumanSeg~\cite{Chu_2022_WACV} (a human body image segmentation method). In summary, our student is labelled with two teachers, 3DDFA-V2 and PP-HumanSeg used for the face and body part teachers, respectively. In our method, a naive way to fuse the two teacher labels is to directly add and average the normalized maps of the face and body part, and then the output is taught to our student model by label knowledge distillation. 
% \vspace{-12pt}
\subsection{Weakly Supervised Loss Functions}
\label{sec:3.4}
Several loss functions are used to train our IFAST. Specifically, weighted relative disparity loss, reconstruction loss, and disparity smooth loss are used to train the depth estimation part. Moreover, focal confidence map loss, confidence map triplet loss, and classification loss \cite{journals/pami/LinGGHD20} are utilized to train the classification part. 
For a better understanding, these various loss functions will be discussed in the following.

\subsubsection{Weighted Relative Disparity Loss} 
In the relative depth label, for each pair of images $L$ and $R$, $K$ pairs of points $p$ in the left image $g_l$ are randomly selected, i.e. $p_k=(i_k, j_k,r_k)$, $k = 1,...,K$, where $i_k=(x_{ik},y_{ik})$ and $j_k = (x_{jk},y_{jk})$ respectively represent the coordinates of the first and second points of $p_k$, and $r_k \in \{+1, -1, 0\}$ represents the relative depth relationship between $i_k$ and $j_k$. For example, if $i_k$ is closer to the camera than $j_k$, the label $r_k$ is $+1$. Conversely, the label $r_k$ is $-1$ when $i_k$ is farther from the camera than $j_k$, and the label $r_k$ is $0$ for other situations. \textbf{Weighted relative disparity loss} is proposed:
\begin{equation}
	\mathcal{L}_w(g,p,d)=\sum_{k=1}^{K} \phi_k(g,i_k,j_k,r_k,d),
\end{equation}
where $d \in \mathbb{R}^{H\times W}$ denotes the disparity map and $\phi_k(\cdot)$ indicates the loss of $k$-th point pair:
\begin{equation}
	\phi_k=\left\{
	\begin{aligned}
		\log(1+\lambda_k\exp(d_{x_{ik},y_{ik}}-d_{x_{jk},y_{jk}})),\ r_k=+1,\\
		\log(1+\lambda_k\exp(d_{x_{jk},y_{jk}}-d_{x_{ik},y_{ik}})),\ r_k=-1,\\
	\end{aligned}
	\right.
\end{equation}
where the weight $\lambda_k$ is calculated from the pseudo-depth in the teacher label $\tilde{d} \in \mathbb{R}^{H\times W}$, since the relative disparity loss should be greater for points with different pseudo-depths:
\begin{equation}
	\lambda_k=\exp(r_k(\tilde{d}_{x_{jk},y_{jk}}-\tilde{d}_{x_{ik},y_{ik}})).
\end{equation}

\subsubsection{Reconstruction Loss}
The ideal disparity map represents the corresponding positional relationship between the pixels of the left and right images, so it is theoretically possible to reconstruct the left image using the disparity map. Reconstruction loss is utilized because training with only relative disparity labels cannot obtain accurate disparity values. According to \cite{wang2020parallax}, the reconstruction loss consists of the $L_1$ loss and SSIM\cite{ssim} loss between the left image and the reconstructed left image.
\begin{equation}
\hat{L}=\mathrm{Rec}(R,d),
\end{equation}
\begin{equation}
\begin{aligned}
\mathcal{L}_r(L,\hat{L})=0.15L_1(L,\hat{L})+0.85(1-\mathrm{SSIM}(L,\hat{L})),
\end{aligned}
\end{equation}
where $L$, $\hat{L}$, $R$, $d$ and $\mathrm{Rec}(\cdot)$ denote the left image, the reconstructed left image, the right image, the disparity map, and the reconstruction operation, respectively.

\subsubsection{Disparity Smooth Loss}
The smooth area in the left image should also be smooth in the corresponding area of its disparity map. Therefore, the disparity smooth loss is selected following \cite{wang2020parallax}. This constraint ensures that the disparity gradients are positively correlated with the gradients of the left image, as follows:
\begin{equation}
\begin{aligned}
&\mathcal{L}_{s}(d,L)=\sum_{i,j}^{H,W} [0.2(|\frac{\partial d_{i,j}}{\partial x}|e^{-|\frac{\partial L_{i,j}}{\partial x}|}+|\frac{\partial d_{i,j}}{\partial y}|e^{-|\frac{\partial L_{i,j}}{\partial y}|})\\&+0.8(|\frac{\partial L_{i,j}}{\partial x}|e^{-|\frac{\partial d_{i,j}}{\partial x}|}+|\frac{\partial L_{i,j}}{\partial y}|e^{-|\frac{\partial d_{i,j}}{\partial y}|})].
\end{aligned}
\end{equation}

\subsubsection{Focal Confidence Map Loss} 
From the dual-teacher distillation framework above, the pseudo-depth labels produced by two teachers can be obtained. These pseudo-depth labels are further leveraged to train our confidence map generator, since their values range from $0$ to $1$. The proposed \textbf{focal confidence map loss} is shown as follows:
\begin{equation}
	\mathcal{L}_f(c,\tilde{d})=\sum_{i,j}^{H,W} e^{(1-logit)}(c_{i,j}-\tilde{d}_{i,j})^{2},
\end{equation}
where $c \in \mathbb{R}^{2\times H\times W}$ is the confidence map produced by the confidence map generator, $\tilde{d} \in \mathbb{R}^{H\times W} $ is the pseudo-depth map taught by the two teachers, and $logit$ indicates the classification logit corresponding to the classification label.

\subsubsection{Confidence Map Triple Loss}
To make better use of disparity to learn the discriminative feature in the confidence map, a \textbf{confidence map triple loss} is proposed in CMG. Specifically, the distances of the latent features among all samples are computed in a batch, and a constraint is imposed that the distances between the features of samples in the same class be smaller than those of samples in different classes:
\begin{equation}
\begin{aligned}
\mathcal{L}_t(F)=\sum_{i=1}^{N} \max(0, dis_{i}^{\mathrm{pos}}-dis_{i}^{\mathrm{neg}} + m),
\end{aligned}
\end{equation}
where $N$ is the batch size, $dis_{i}^{\mathrm{pos}}$ is the maximum distance between the $i$-th sample and the samples of the same class, $dis_{i}^{\mathrm{neg}}$ is the minimum distance between the $i$-th sample and the samples of the different classes, and $m$ denotes a hyperparameter indicating the margin.

\section{Experiments}
\subsection{Introduction to the BNI-FAS Dataset}
\begin{figure}[t]
    \centering
	%\begin{spacing}{0.7}
	\begin{minipage}[t]{0.18\linewidth}
		\centerline{\includegraphics[width=\linewidth]{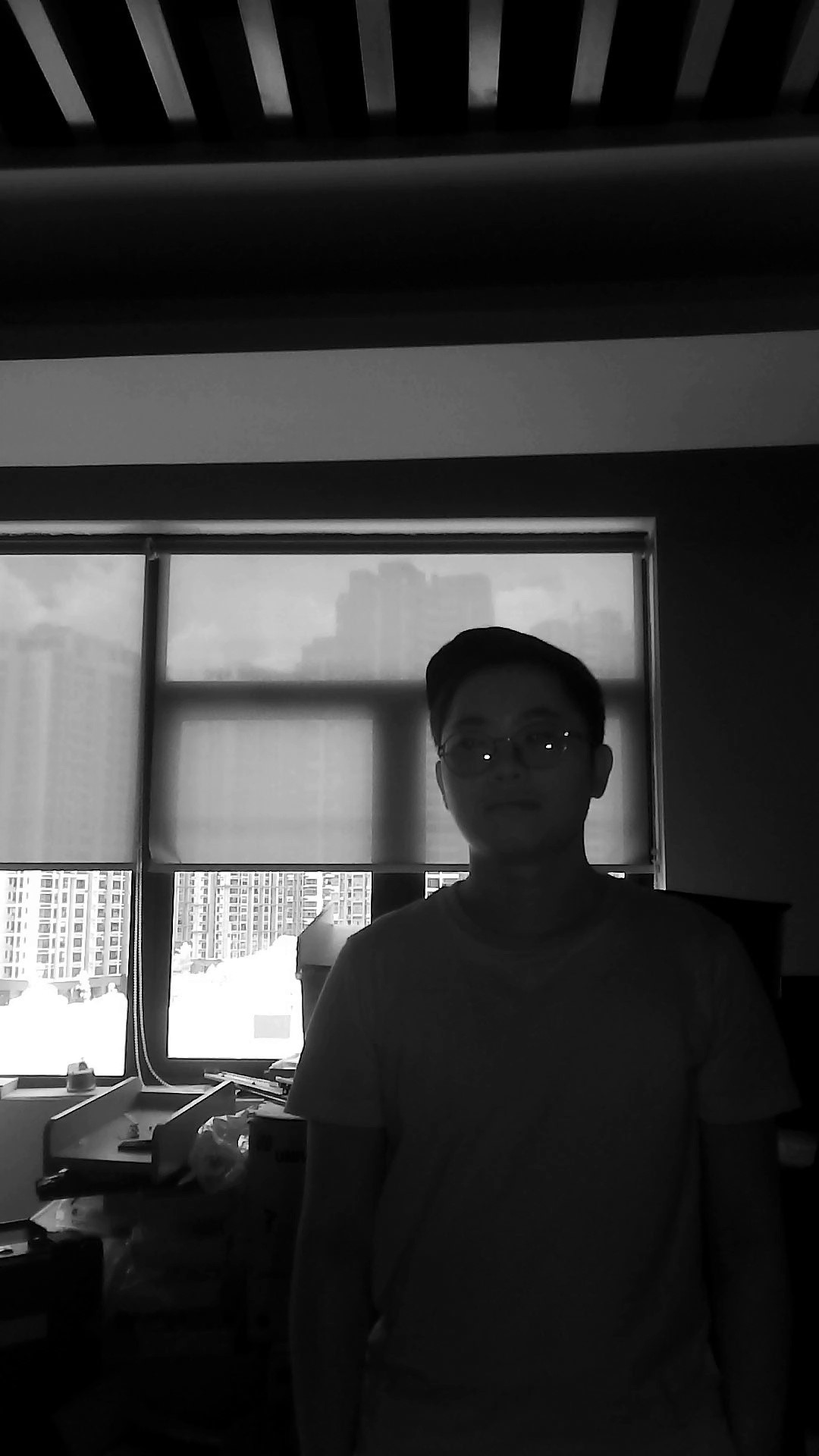}}
	\end{minipage}
	\begin{minipage}[t]{0.18\linewidth}
		\centerline{\includegraphics[width=\linewidth]{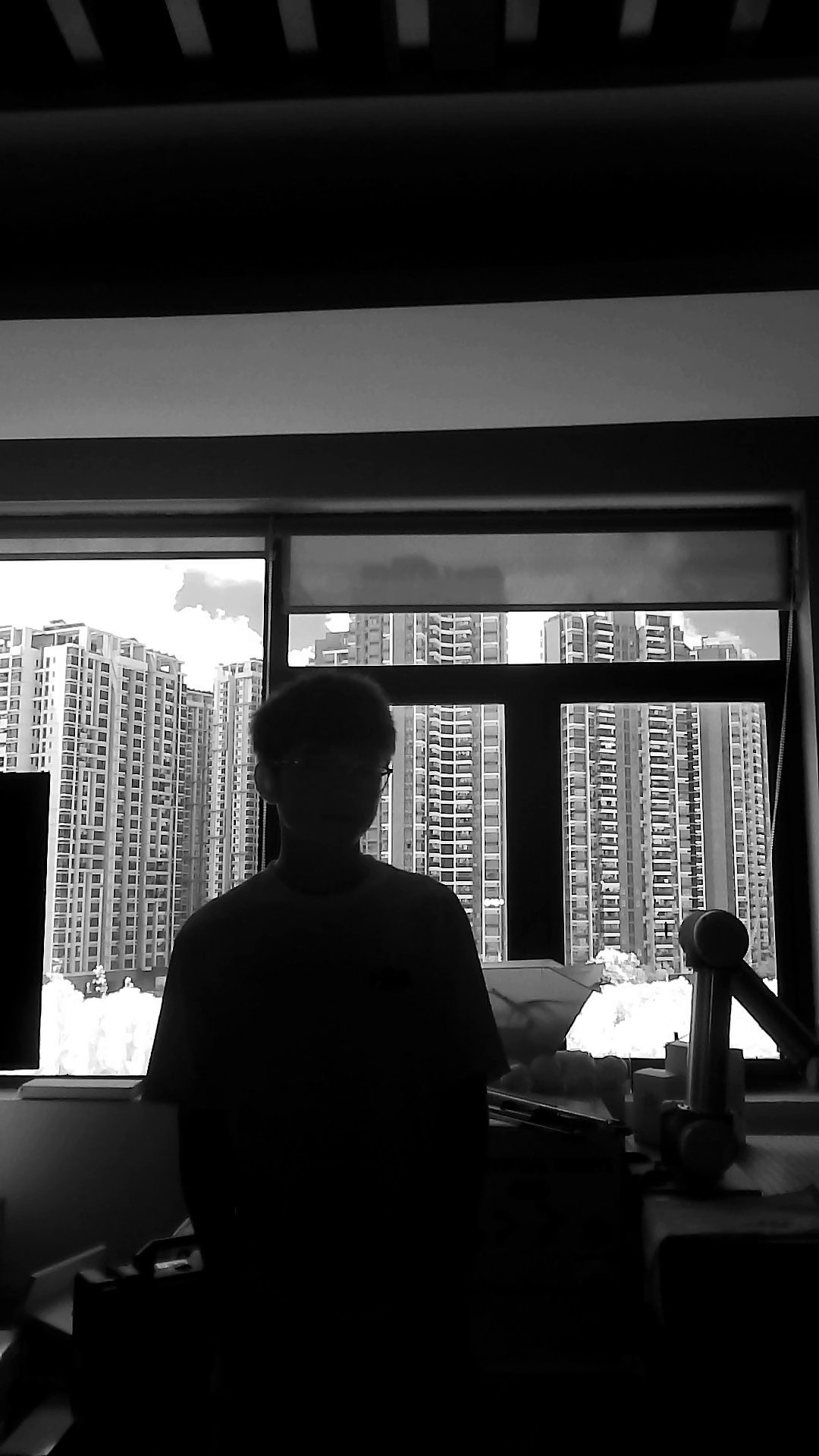}}
	\end{minipage}
	\begin{minipage}[t]{0.18\linewidth}
		\centerline{\includegraphics[width=\linewidth]{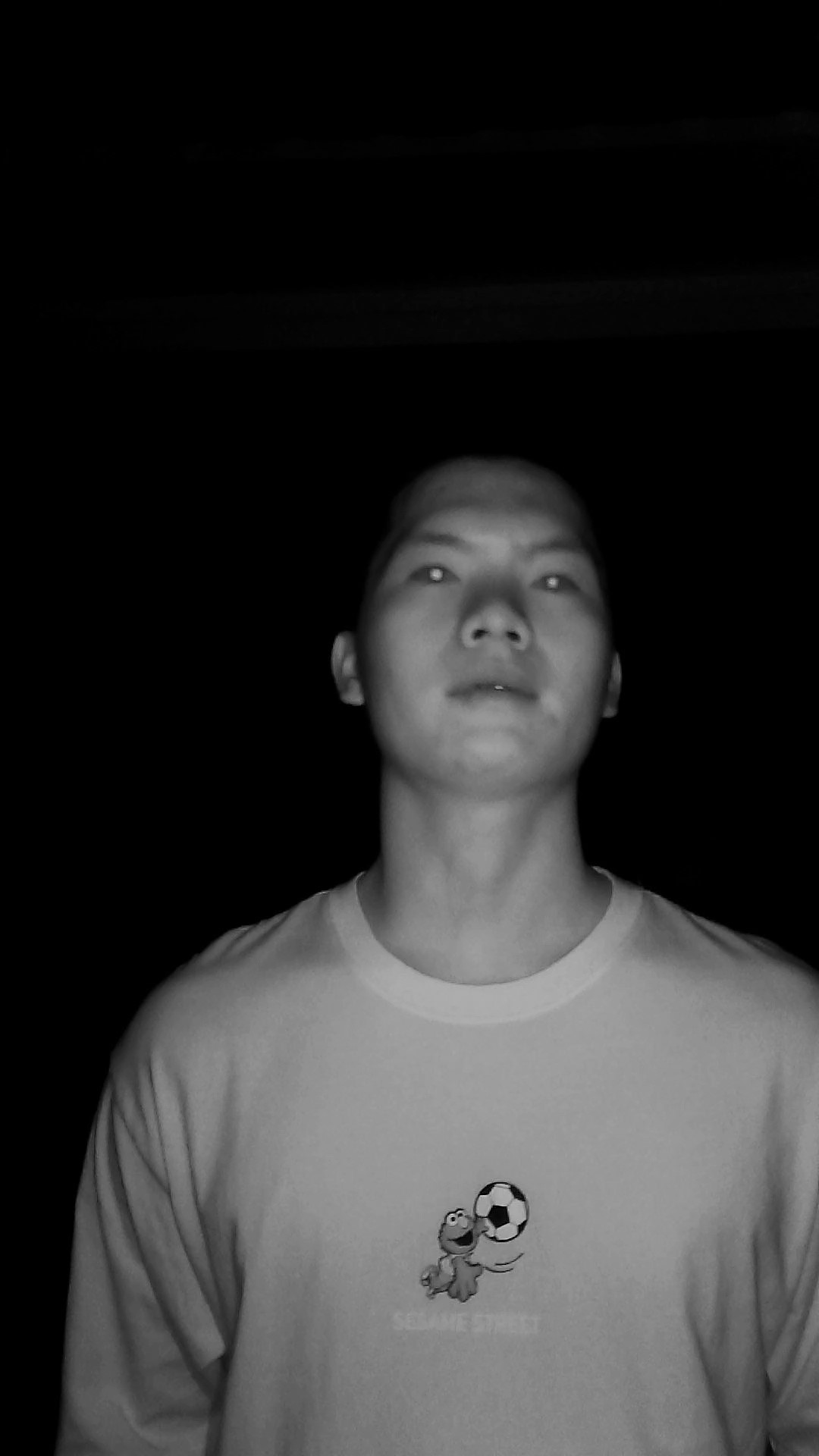}}
	\end{minipage}
    \begin{minipage}[t]{0.18\linewidth}
		\centerline{\includegraphics[width=\linewidth]{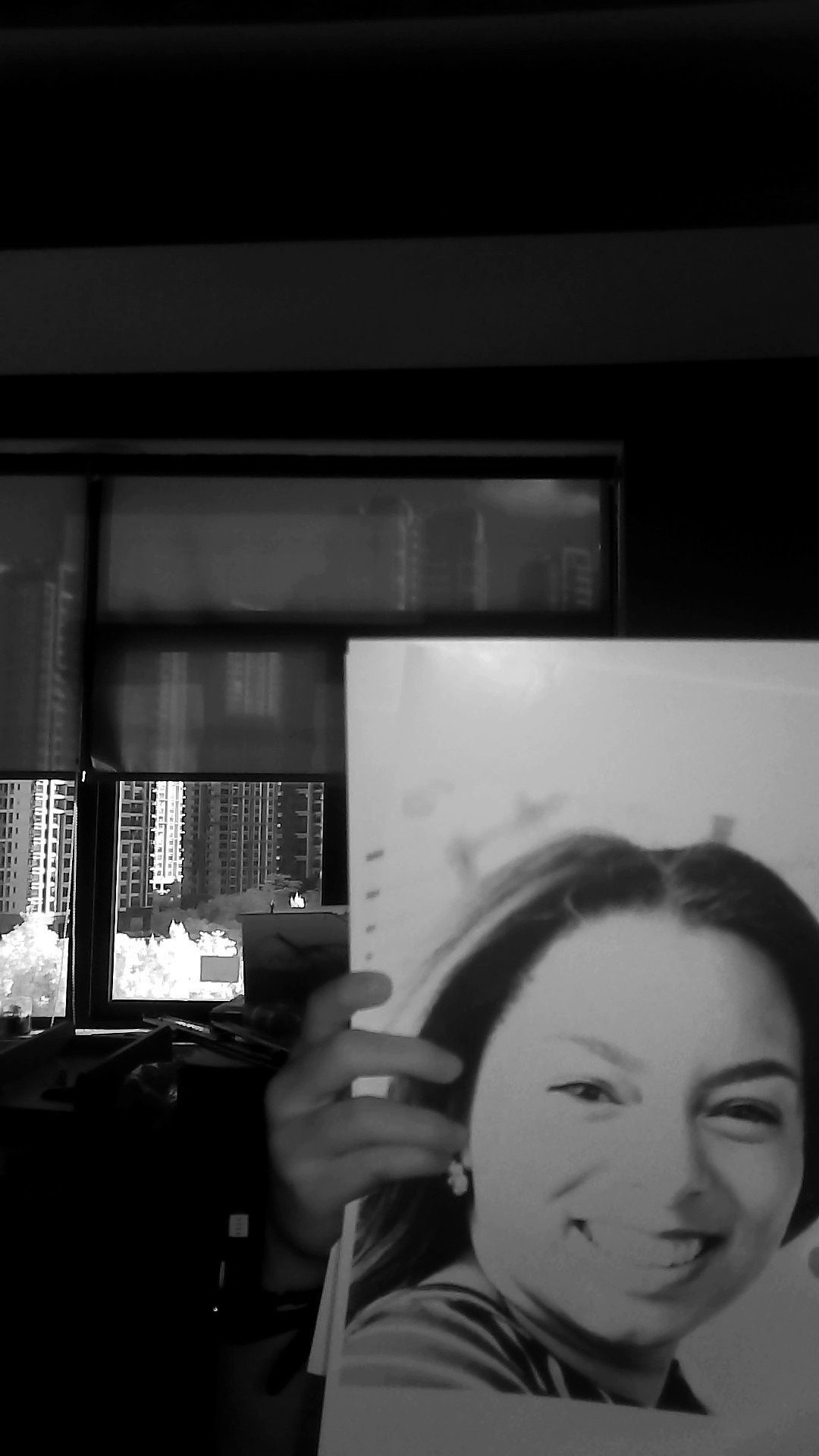}}
	\end{minipage}
    \begin{minipage}[t]{0.18\linewidth}
		\centerline{\includegraphics[width=\linewidth]{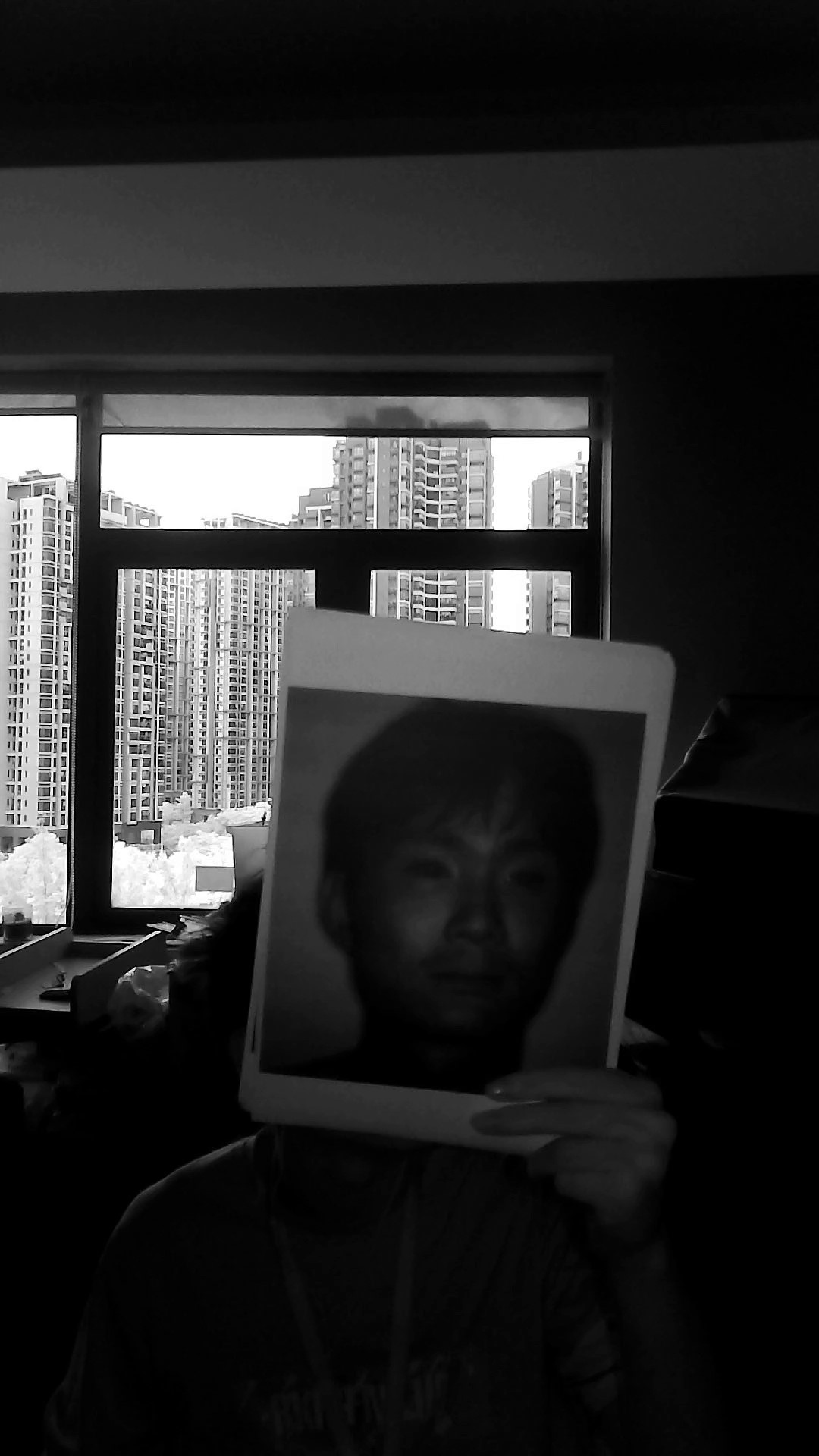}}
	\end{minipage}
        % \vspace{-8pt}
	\caption{Examples of our BNI-FAS.}
	\label{examplefig}
    % \vspace{-10pt}
\end{figure}

A large number of real-face and plane-attack images are captured using industrial-customized binocular near-infrared (NIR) cameras, and a large-scale binocular near-infrared (NIR) image dataset called BNI-FAS is constructed. Due to the inherent reflective properties of NIR cameras when shooting at the screens, NIR cameras have better face anti-spoofing (FAS) performance against screen attacks. Hence, our dataset only contains images printed on paper as plane attacks. Besides, our dataset focuses on paper-based attacks because they are the most common attacks other than screen-based attacks. Considering that the 3D facial model attacks are costly and rare, our dataset omits the acquisition of 3D facial models. Remarkably, our dataset especially emphasizes the effect of different illumination, distances, and paper textures for FAS. Specifically, our binocular NIR image dataset is divided into two sessions, which are taken at different locations. They contain a total of about 340,000 pairs of binocular images, where 146,400 pairs are real face images and 190,000 pairs are plane attacks. More specifically, there are 90,000 pairs of images (60,000 pairs of real faces and 30,000 pairs of plane attacks) in session 1. Session 2 has 250,000 pairs of images, of which approximately 86,400 are real faces and 160,000 are plane attack images.
The main advantage of our BNI-FAS is that various shooting conditions are controlled, including 5 illumination conditions, 4 distances, 8 paper materials, and the presence of a white box. Some example images are shown in Fig. \ref{examplefig}. The detailed settings are presented in Table \ref{BNI-FAStab}.

\begin{table}[t]
    \centering
	\resizebox{0.8\linewidth}{!}{
	\begin{tabular}{m{1.3cm}m{1.3cm}m{1.3cm}m{1.3cm}m{1.3cm}}
		\noalign{\smallskip}\hline\noalign{\smallskip}
		\multicolumn{1}{c}{\multirow{2}*{\textbf{Settings}}}
		&\multicolumn{2}{c}{Session1}&\multicolumn{2}{c}{Session2}\\\cmidrule(l){2-3}\cmidrule(l){4-5}
		&Real&Attack&Real&Attack\\		
		\noalign{\smallskip}\hline\noalign{\smallskip}
		ID&6&None&4&None\\
		\noalign{\smallskip}\hline\noalign{\smallskip}
		\multirow{5}*{\tabincell{l}{\\\\Illumination\\\\}}
		&\multicolumn{2}{c}{\multirow{3}*{\tabincell{l}{Night-Light\\Weak-Light\\Full-Light}}}&\multicolumn{2}{c}{\multirow{5}*{\tabincell{l}{Night-Light\\ Low-Light\\ Weak-Light\\ Normal-Light\\ Full-Light}}}\\\\\\\\\\
		\noalign{\smallskip}\hline\noalign{\smallskip}
		\multirow{4}*{\tabincell{l}{\\Distance\\\\}}
		&\multicolumn{2}{c}{\multirow{3}*{\tabincell{l}{0.5m\\1m\\1.5m}}}&\multicolumn{2}{c}{\multirow{4}*{\tabincell{l}{0.5m\\1m\\1.5m\\2m}}}\\\\\\\\
		\noalign{\smallskip}\hline\noalign{\smallskip}
		\multirow{2}*{\tabincell{l}{White Box\\}}
		&\multicolumn{2}{c}{\multirow{2}*{\tabincell{l}{Yes\\No}}}&\multicolumn{2}{c}{\multirow{1}*{\tabincell{l}{No}}}\\\\
		\noalign{\smallskip}\hline\noalign{\smallskip}
		
		\multirow{8}*{\tabincell{l}{\\\\Paper\\Material\\\\}}
		&\multicolumn{2}{c}{\multirow{3}*{\tabincell{l}{RGB-A4\\Black-A4\\NIR-A4}}}&\multicolumn{2}{c}{\multirow{8}*{\tabincell{l}{RGB-A4\\RGB-IceWhite\\RGB-Coated\\ NIR-A4\\ NIR-Coated\\ NIR-IceWhite\\
					Black-A4\\Black-NIR}}}\\\\\\\\\\\\\\\\
		\noalign{\smallskip}\hline\noalign{\smallskip}
	\end{tabular}
	}
        \vspace{5pt}
	\caption{Details of our dataset. Images of real face and plane attack share the same settings of illumination and distance in 2 sessions. }
    % \vspace{-20pt}
    \label{BNI-FAStab}
\end{table}
% \vspace{-4pt}
\subsection{Implementation Details}\label{Imp de}
IFAST can be divided into two parts and trained sequentially, where the disparity estimation Transformer is trained first followed by the CMG. The left and right images from our NIR camera have a native resolution of $1920\times 1080$ pixels. The face detector, same as \cite{conf/eccv/GuoZYYLL20}, is first used to obtain the position of the face in the image and crop the face part, and then resized to $256\times 256$. The depth estimation part is initialized with pre-trained parameters on Sceneflow \cite{MIFDB16} and then trained with our NIR binocular dataset with a learning rate starting of $0.0001$. The Adam optimizer is adopted to train the depth estimation network for about $1600$ steps. Weighted relative disparity loss, reconstruction loss, and disparity smooth loss are used to train the depth estimation part. For the classification part, focal confidence map loss, confidence map triplet loss, and classification loss \cite{journals/pami/LinGGHD20} are adopted to train it. The weights of all loss terms are set to $1$ during the experiment. The initial weights of the classification part are set randomly, the Adam optimizer is used to train the classification part for about $3000$ steps with the batch size of $50$.

\begin{table*}[t]
	\centering
	\resizebox{\linewidth}{!}{
		\begin{tabular}{cccccccccc}
			\hline
			Method & ACC$\uparrow$ & AUC$\uparrow$ & EER$\downarrow$ & TPR(1.\%)$\uparrow$ & TPR(0.5\%)$\uparrow$ &TPR(0.1\%)$\uparrow$ &TPR(0.05\%)$\uparrow$ &TPR(0.01\%)$\uparrow$ &TPR(0.001\%)$\uparrow$\\
			\hline
			PSMNet$^*$  & 94.94\% &0.98920 & 4.731\% & 84.43\% & 76.38\% & 50.51\% & 42.58\% & 23.43\% & 13.92\%\\
			StereoNet$^*$  & 95.87\% &0.99618 & 2.398\% & 87.15\% & 78.48\% & 59.88\% & 55.27\% & 8.00\% &0\% \\
			GwcNet$^*$  & 97.80\% &0.99743 & 2.538\% & 95.93\% & 94.20\% & 84.01\% & 71.32\% &12.21\% & 0\%\\
			STTR$^*$  & 86.34\% &0.98594 & 7.159\% & 83.98\% & 81.30\% & 75.39\% & 71.78\% &60.60\% & 50.22\%\\
			PASMNet  & 97.14\% &0.99710 & 1.852\% & 96.65\% & 93.95\% & 81.22\% & 73.55\% & 58.38\% &30.49\%\\\
			Dual-Net$^*$ & \underline{98.69\%} &0.99240 & 1.438\% & 98.12\% & 97.57\% & \underline{91.76\%} & \underline{85.26\%} &\underline{76.23\%} &62.46\% \\
			BM & 96.58\% &\underline{0.99954} & \underline{0.811\%} & \underline{99.40\%} & \underline{98.39\%} & 90.16\% & 84.10\% &73.71\% &\underline{66.64\%} \\
			SGBM & 92.77\% &0.99857 & 1.381\% & 97.56\% & 93.16\% & 72.26\% & 61.75\% &38.01\% &13.21\% \\
            IFAST & \textbf{99.41}\% & \textbf{0.99990}& \textbf{0.3657\%} & \textbf{99.85\%} & \textbf{99.75\%} & \textbf{98.18\%} & \textbf{96.70\%} & \textbf{91.52\%} & \textbf{79.74\%}\\
			\hline
		\end{tabular}
	}
    % \vspace{-1pt}
    \vspace{5pt}
	\caption{Results of different depth estimation methods with our confidence map generator on the BNI-FAS dataset. The best and second best are in bold and underlined, respectively. The sign $*$ denotes that it is an full-supervised method.}
	\label{comp_depth}
    % \vspace{-12pt}
\end{table*}
 \begin{figure*}[t]
	\begin{center}
		\includegraphics[width=\linewidth]{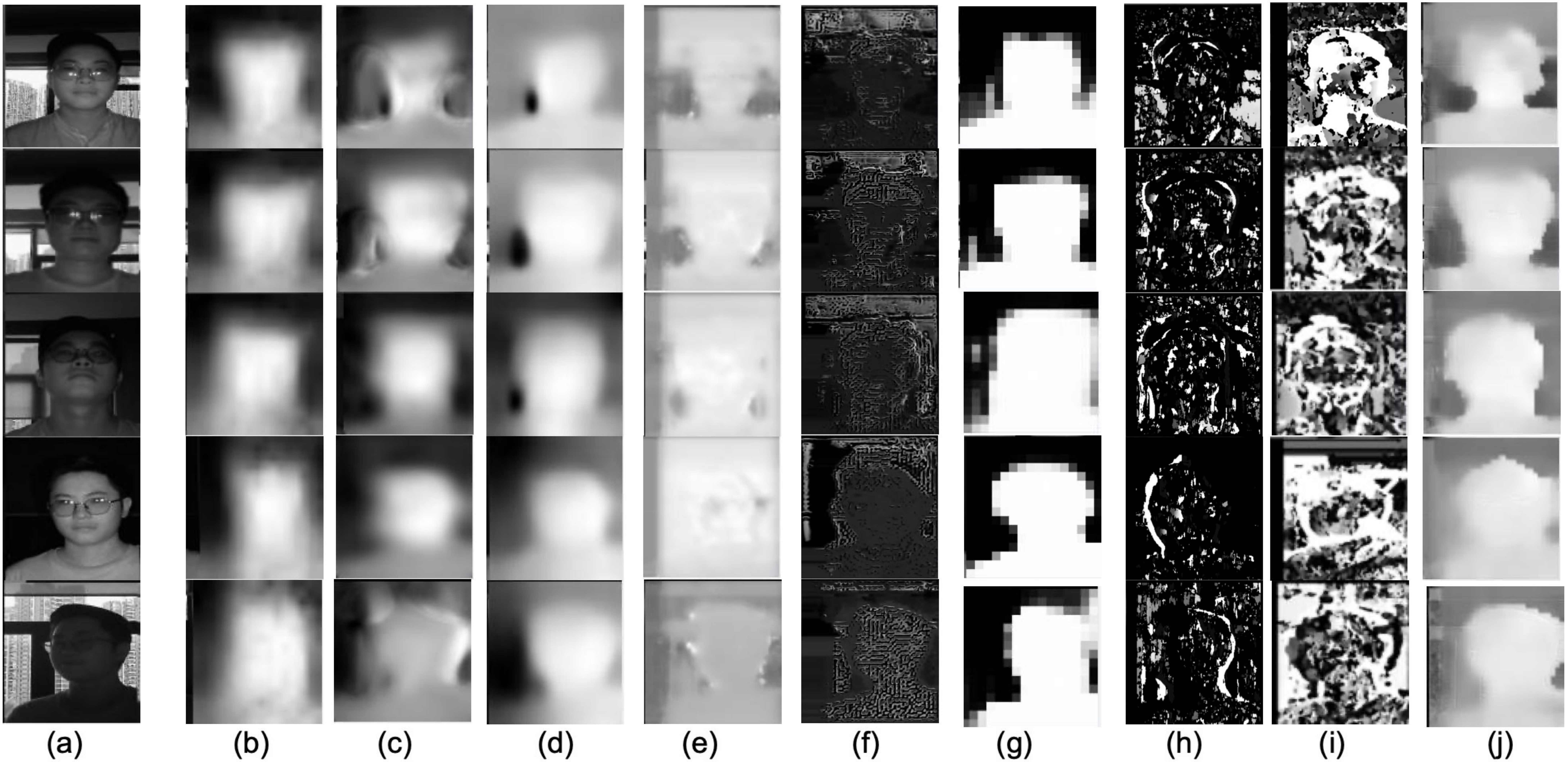}
	\end{center}
        % \vspace{-15pt}
	\caption{The visual comparison of the disparity maps predicted by different methods for estimating the depth of real faces. (a) The left image. (b) StereoNet. (c) PSMNet. (d) GwcNet. (e) STTR. (f) PASMNet. (g) Dual-Net. (h) BM. (i) SGBM. (j) IFAST. Our IFAST yields higher visibility.}
	\label{real}
    % \vspace{-10pt}
\end{figure*}
\begin{figure}[t]
	%\hspace{0.5cm}
	\centering
	\includegraphics[width=0.8\linewidth]{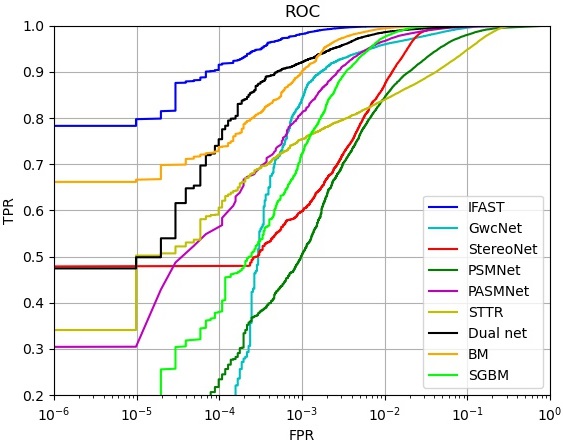}
        % \vspace{-8pt}
	\caption{The classification ROC curves of different depth estimation methods. Given an FPR, the higher the TPR, the better the performance. }
	\label{roc}
    % \vspace{-10pt}
\end{figure}

% \vspace{-6pt}
\subsection{Comparison of Different Depth Estimation Methods}

The basic purpose of our disparity estimation Transformer for generating depth maps is for binary classification and therefore does not require the pursuit of excessive stereo matching performance. Moreover, BNI-FAS is collected in real scenes and reflects irregularities. This phenomenon is different from most stereo matching datasets \cite{MIFDB16}. We select the following compared methods, including traditional depth estimation methods BM \cite{bm}, SGBM \cite{SGBM}, deep-network-based stereo matching methods PSMNet \cite{ChangC18}, StereoNet \cite{KhamisFRKVI18}, GwcNet \cite{GuoYYWL19}, STTR \cite{journals/corr/abs-2011-02910}, PASMNet\cite{wang2020parallax} and depth-based dual-pixel FAS method (Dual-Net) \cite{wudual}. These models are trained on our dataset by fitting their released pre-trained models on our dataset and FAS task. \textcolor{black}{As binocular depth-based experiments require depth estimation from stereo images, only binocular depth-based methods are compared in the experiments.}

To compare the performance of different depth estimation methods on FAS, the proposed confidence map generator is applied to each compared depth estimation method to ensure that the difference in the FAS performance of the compared methods is only related to their depth estimation networks. The Accuracy (ACC), Area Under the Curve (AUC), Equal Error Rate (EER), and True Positive Rate (TPR) at a fixed False Positive Rate (FPR) are calculated for all methods. The results are reported in Table~\ref{comp_depth} and Fig.~\ref{roc}. Using our confidence map generator, all methods achieve satisfactory FAS performance. Compared with other stereo matching depth estimation methods, the proposed IFAST achieves a better FAS performance on metrics such as TPR and ACC. In particular, when the FPR equals to 0.1\%, the TPR metrics of most of the compared methods decrease to less than 0.9, while our IFAST still maintains a relatively good performance. Notably, considerable performance was also achieved by Dual-Net \cite{wudual}. Our dataset emphasizes the use of disparity for FAS, resulting in more reliable and interpretable results in real scenes. 

\begin{figure*}[t]
	\begin{center}
		\includegraphics[width=\linewidth]{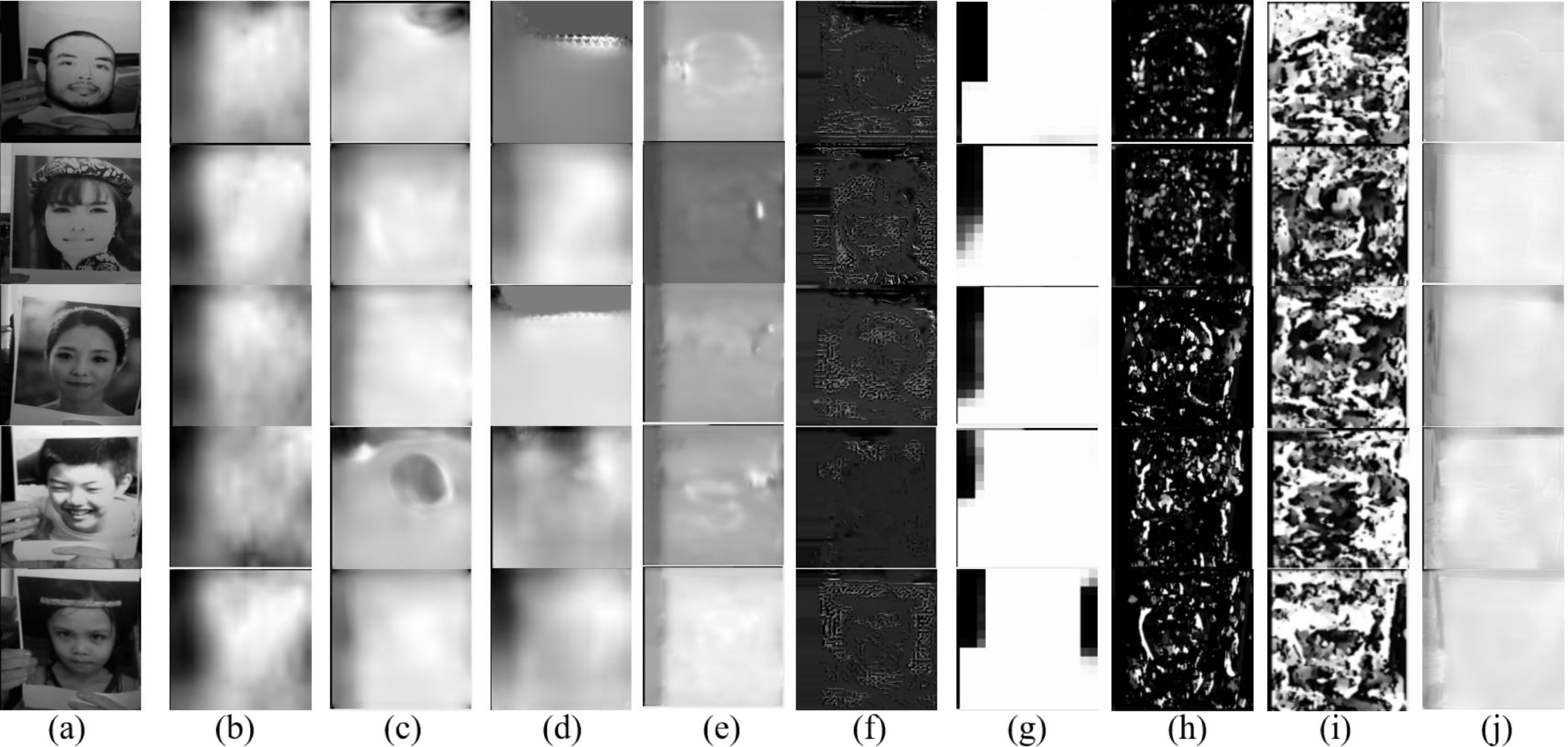}
	\end{center}
    % \vspace{-14pt}
	\caption{The visual comparison of the disparity maps predicted by different methods for estimating the depth of plane attacks. (a) The left image. (b) StereoNet. (c) PSMNet. (d) GwcNet. (e) STTR. (f) PASMNet. (g) Dual-Net. (h) BM. (i) SGBM. (j) IFAST. Our IFAST yields higher visibility.}
	\label{attack}
\end{figure*}
\begin{table*}[t]
	\centering
	\resizebox{\linewidth}{!}{
		\begin{tabular}{cccccccccccc}
			\hline
			Method & Real  & Attack & ACC$\uparrow$ & AUC$\uparrow$ & EER$\downarrow$ & TPR(1.\%)$\uparrow$ & TPR(0.5\%)$\uparrow$ &TPR(0.1\%)$\uparrow$ &TPR(0.05\%)$\uparrow$ &TPR(0.01\%)$\uparrow$ &TPR(0.001\%)$\uparrow$\\
			\hline
			IFAST & S2-W & NIR-A4 & 98.59\% & 0.99478& 1.179\% & 98.79\% & 98.63\% & 97.96\% & 97.44\%& 86.07\% & 34.12\%\\
			IFAST & S2-W & RGB-Coated & 98.63\% & 0.99870& 1.139\% & 98.83\% & 98.65\% & 96.87\% & 92.90\%& 80.76\% & 55.25\%\\
			IFAST & S2-W & RGB-A4 & 99.16\% & 0.99920& 1.0376\% & 98.95\% & 98.82\% & 98.30\% & 97.87\%& 85.38\% & 78.30\%\\
			IFAST & S2-W & Black-NIR & 98.79\% & 0.99890& 1.0579\% & 98.92\% & 98.78\% & 98.22\% & 97.85\% & 93.55\% & 92.81\%\\
			IFAST & S2-W & RGB-Ice & 98.62\% & 0.99869& 1.0579\% & 98.92\% & 98.82\% & 98.48\% & 98.06\% & 77.69\% & 48.26\%\\
			IFAST & S2-W & NIR-Coated & 98.53\% & 0.99433& 1.194\% & 98.78\% & 98.63\% & 97.77\% & 97.17\% & 69.55\% & 55.67\%\\
			\hline
	\end{tabular}}
    % \vspace{-3pt}
    \vspace{5pt}
	\caption{Comparison of paper attacks with different materials. Real and Attack denote the setting of positive samples and negative samples in the test set, respectively.}
	\label{papermaters}
    % \vspace{-20pt}
\end{table*}

\begin{table*}[t]
	\centering
	\resizebox{\linewidth}{!}{
		\begin{tabular}{cccccccccccc}
			\hline
			Method & Real  & Attack & ACC$\uparrow$ & AUC$\uparrow$ & EER$\downarrow$ & TPR(1.\%)$\uparrow$ & TPR(0.5\%)$\uparrow$ &TPR(0.1\%)$\uparrow$ &TPR(0.05\%)$\uparrow$ &TPR(0.01\%)$\uparrow$ &TPR(0.001\%)$\uparrow$\\
			\hline
            FASNet$*$ & S1-Y & NIR-A4 & 85.26\% & 0.69240& 37.11\% & 53.15\% & 52.96\% & 50.00\% & 45.89\%& 30.00\% & 6.30\% \\
            LivenesSlight$*$ & S1-Y & NIR-A4 & \underline{91.42\%} & 0.88680& 5.06\% & 73.97\% & 0.0\% & 0.0\% & 0.0\%& 0.0\% & 0.0\%\\
            CDCN$*$ & S1-Y & NIR-A4 & 60.11\% & \underline{0.99144} & \underline{4.947\%} & 76.57\% & 72.75\% & 63.93\% & 62.01\% & 56.72\% & 56.72\%\\
            DC-CDN$*$ & S1-Y & NIR-A4 & 44.61\% & 0.99061 & 5.58\% & \underline{80.26\%} & \underline{75.09\%} & \underline{66.15\%} & \underline{64.64\%} & \underline{60.03\%} & \underline{60.03\%}\\
            ViTranZFAS$*$ & S1-Y & NIR-A4 & 74.75\% & 0.94670& 13.32\% & 38.67\% & 34.36\% & 26.95\% & 24.25\%& 14.71\% & 10.50\%\\
            SA-FAS$*$ & S1-Y & NIR-A4 & 86.45\% & 0.944664& 13.9732\% & 40.38\% & 39.35\% & 37.92\% & 36.95\%& 33.81\% & 18.92\%\\
            Entry-V2$*$ & S1-Y & NIR-A4 & 73.26\% & 0.829160& 25.60\% & 4.15\% & 1.62\% & 0.308\% & 0\%& 0\% & 0\%\\
            PatchNet$*$ & S1-Y & NIR-A4 & 82.46\% & 0.745312& 33.80\% & 45.33\% & 33.79\% & 0\% & 0\%& 0\% & 0\%\\
			IFAST &S1-Y & NIR-A4 & \textbf{94.37\%} & \textbf{0.99336}& \textbf{4.165\%} & \textbf{90.34\%} & \textbf{87.15\%} & \textbf{75.69\%} & \textbf{72.29\%} & \textbf{61.96\%} & \textbf{46.10\%}\\
			\hline
			FASNet$*$ & S1-Weak & RGB-A4 & 83.85\% & 0.85480& 25.45\% & 54.74\% & 53.38\% & 50.78\% & 49.11\% & 43.17\% & 19.51\%\\
            LivenesSlight$*$ & S1-Weak & RGB-A4 & 85.28\% & 0.81810& 10.23\% & 0.0\% & 0.0\% & 0.0\% & 0.0\%& 0.0\% & 0.0\%\\
            CDCN$*$ & S1-Weak & RGB-A4 & 79.49\% & 0.9458& 12.01\% & 28.98\% &19.35\% & 9.48\% & 6.64\%& 4.01\% & 2.85\%\\
            DC-CDN$*$ & S1-Weak & RGB-A4 & 80.32\% & \underline{0.9953} & \underline{2.623\%} & \underline{94.88\%} &\underline{91.96\%} & \underline{80.48\%} & \underline{75.51\%}& \underline{68.08\%} & \underline{47.99\%}\\
            ViTranZFAS$*$ & S1-Weak & RGB-A4 & \underline{96.33\%} & 0.99150& 3.722\% & 84.83\% & 66.39\% & 13.67\% & 2.00\%& 0.0\% & 0.0\%\\
            SA-FAS$*$ & S1-Weak & RGB-A4 & 83.49\% & 0.937388& 14.73\% & 37.81\% & 29.74\% & 20.46\% & 18.50\%& 13.12\% & 10.38\%\\
            Entry-V2$*$ & S1-Weak & RGB-A4 & 85.29\% & 0.918709& 16.94\% & 33.11\% & 26.43\% & 17.61\% & 14.97\%& 7.17\% & 0\%\\
            PatchNet$*$ & S1-Weak & RGB-A4 & 84.82\% & 0.745517& 38.83\% & 49.74\% & 49.39\% & 46.79\% & 40.44\%& 0\% & 0\%\\
            IFAST & S1-Weak & RGB-A4 & \textbf{95.73\%} & \textbf{0.99740}& \textbf{1.329\%} & \textbf{98.42\%} & \textbf{97.30\%} & \textbf{89.14\%} & \textbf{83.94\%} & \textbf{71.54\%} & \textbf{49.88\%}\\
			\hline
			FASNet$*$ & S2-Night & RGB-A4 & 87.40\% & 0.97840& 9.248\% & 81.70\% & 80.40\% & \underline{77.58\%} & \underline{74.42\%} & 59.94\% & 13.18\% \\
            LivenesSlight$*$ & S2-Night & RGB-A4 & 89.91\% & 0.89790& 8.103\% & 0.0\% & 0.0\% & 0.0\% & 0.0\%& 0.0\% & 0.0\%\\
            CDCN$*$ & S2-Night & RGB-A4 & 91.11\% & 0.9860& 6.11\% & 74.10\% & 68.64\% & 61.11\% & 56.59\%& 46.84\% & 38.53\%\\
            DC-CDN$*$ & S2-Night & RGB-A4 & 76.46\% & \underline{0.99832} & \underline{1.542\%} & \underline{95.92\%} & \underline{88.68\%} & 72.03\% & 66.99\%& \underline{61.68\%} & \underline{47.75}\%\\
            ViTranZFAS$*$ & S2-Night & RGB-A4 & \underline{97.12\%} & 0.99490 & 2.505\% &91.48\% & 75.71\% & 6.01\% & 0.73\%& 0.0\% & 0.0\%\\
            SA-FAS$*$ & S2-Night & RGB-A4 & 94.17\% & 0.989711& 4.671\% & 79.62\% & 70.48\% & 31.34\% & 18.65\%& 5.784\% & 3.285\% \\
            Entry-V2$*$ & S2-Night & RGB-A4 & 89.75\% & 0.963851& 10.1602\% & 48.22\% & 33.89\% & 11.94\% & 8.02\%& 0\% & 0\%\\
            PatchNet$*$ & S2-Night & RGB-A4 & 82.32\% & 0.82895& 16.375\% & 58.40\% & 55.69\% & 43.78\% & 33.28\%& 0\% & 0\%\\
            IFAST & S2-Night & RGB-A4 & \textbf{98.81\%} &\textbf{ 0.99990}& \textbf{0.186\%} & \textbf{100\%} & \textbf{99.98\%} & \textbf{98.79\%} & \textbf{95.57\%} & \textbf{74.49\%} & \textbf{52.94\%}\\
			\hline
	\end{tabular}}
    % \vspace{-3pt}
    \vspace{5pt}
	\caption{Comparison of FAS methods on different test settings. Real and Attack denote the setting of positive samples and negative samples in the test set, respectively. The best and second best are in bold and underlined, respectively. The sign $*$ denotes that it is an full-supervised method.}
	\label{changeset}
    % \vspace{-13pt}
\end{table*}

% \vspace{-8pt}

\subsection{\textcolor{black}{Comparison of Generalization Performance with Different FAS Methods}}

\textcolor{black}{In order to investigate the generalization performance of IFAST under different conditions, some single-shot monocular FAS methods are selected such as FASNet \cite{LucenaJMSVL17}, LivenesSlight \cite{9130568}, CDCN \cite{conf/cvpr/YuZWQ0LZZ20}, DC-CDN\cite{DCCDN}, ViTranZFAS \cite{journals/corr/abs-2011-08019}, SA-FAS~\cite{sun2023rethinking}, Entry-V2~\cite{sergievskiy2022generalizable} and PatchNet~\cite{wang2022patchnet} for comparison. Our training set is fixed and the test set is selectively changed to evaluate performance under specific conditions.
As can be seen from Table \ref{changeset}, when chooseing the night-light images of other IDs in session 2 as the test set, IFAST performs better than the monocular methods. The weakened texture features of the night-light images also have a huge impact on our depth estimation Transformer in terms of stereo matching. Similarly, the images of ID Y in session 1 or the weak-light images in session 1 without a white box are set as the positive samples in the test set, respectively. As can be observed in Table \ref{changeset}, IFAST also achieves better performance than monocular methods, because monocular methods are texture-based and rely too much on texture features to discriminate, while some features may depend on specific scenes or conditions. As a result, poor performance of monocular FAS methods arises when the conditions of the test set change significantly from those of the training set. 
% To verify the robustness of IFAST against paper attacks with different materials,
As can be seen in Table \ref{papermaters}, IFAST can achieve superior performance for different paper materials, even though these types of papers are very different from the Black-A4 used in the training set. This is because IFAST is based on binocular depth estimation, and the depth maps of papers with different materials should theoretically be the same, making IFAST cope with different paper attacks.} 

\begin{table*}[t]
	\centering
	\resizebox{\linewidth}{!}{
		\begin{tabular}{cccccccccc}
			\hline
			Setting & ACC$\uparrow$ & AUC$\uparrow$ & EER$\downarrow$ & TPR(1.\%)$\uparrow$ & TPR(0.5\%)$\uparrow$ &TPR(0.1\%)$\uparrow$ &TPR(0.05\%)$\uparrow$ &TPR(0.01\%)$\uparrow$ &TPR(0.001\%)$\uparrow$\\
			\hline
            Ablation study on losses\\
            \hline
			W/o relative disparity loss & 98.59\% &0.99923 & 1.097\% & 98.70\% & 96.98\% & 85.54\% & 80.06\% & 65.79\% &51.86\%\\
			W/o reconstruction loss & \underline{99.20\%} &0.99969 & 0.5379\% & 99.79\% & 99.39\% & 94.18\% & 91.78\% & 78.42\% &11.94\%\\
			W/o disparity smooth loss & 98.78\% &0.99930 & 0.9664\% & 99.07\% & 97.69\% & 84.79\% & 73.80\% & 48.52\% &26.49\%\\
			W/o focal confidence map loss& 90.41\% &0.99423 & 3.173\% & 93.67\% & 90.45\% & 75.31\% & 68.06\% &57.03\% & 33.81\%\\
			W/o confidence map triplet loss & 98.30\% &\underline{0.99990} & \underline{0.3758\%} & \underline{99.82\%} & \underline{99.72\%} & \textbf{99.01\%} & \textbf{96.79\%} &\underline{90.92\%} & 68.51\%\\
			\midrule
            Ablation study on disparity estimation Transformer&\\
            \hline
			W/o up/down sampling & 94.06\% &0.99794 & 1.766\% & 97.91\% & 97.22\% & 93.94\% & 91.44\% &80.10\% & 66.63\%\\
			W/o DMA block & 98.47\% &0.99924 & 1.002\% & 98.99\% & 98.57\% & 96.52\% & 93.58\% &85.99\% & 71.97\%\\
			\midrule
            Ablation study on confidence map generator (CMG)&\\
            \hline
			W/o gated in CMG & 98.85\% &0.99950 & 0.8130\% & 99.38\% & 98.67\% & 93.76\% & 84.98\% &41.28\% &9.00\% \\
			Only feed disparity in CMG & 98.32\% &0.99964 & 0.8434\% & 99.29\% & 98.72\% & 96.85\% & 95.22\% &89.13\% &\textbf{82.58\%} \\
            \midrule
			Full IFAST & \textbf{99.41\%} & \textbf{0.99990}& \textbf{0.3657\%} & \textbf{99.85\%} & \textbf{99.75\%} & \underline{98.18\%} & \underline{96.70\%} & \textbf{91.52\%} & \underline{79.74\%} \\
            \hline
	\end{tabular}}
    % \vspace{-1pt}
    \vspace{5pt}
	\caption{Ablation studies of IFAST. W/o is an abbreviation for 'without', which indicates the removal of a component or supervision. The best and second best are in bold and underlined, respectively. }
	\label{abs}
    % \vspace{-20pt}
\end{table*}

\subsection{Ablation Studies}\label{sec:abs}
\subsubsection{Analysis of Estimated Disparity}
\textcolor{black}{In addition to comparing the performance of FAS, we also compare the depth maps obtained by different methods to emphasize that the use of depth information for FAS is interpretable. The depth maps of the real face and the plane attack are shown in Fig.~\ref{real} and Fig.~\ref{attack}, respectively. PSMNet \cite{ChangC18} achieves mediocre results with uneven potholes, while the depth maps estimated by StereoNet \cite{KhamisFRKVI18} are not smooth. GwcNet \cite{GuoYYWL19} visually achieves relatively good results. The results of STTR \cite{journals/corr/abs-2011-02910} are not discriminative enough due to the poorly estimated depth of the background, resulting in poor FAS performance. PASMNet \cite{wang2020parallax} is a completely self-supervised method and relies on photometric reconstruction, resulting in an inaccurate depth map with a lot of noise in our real NIR images. Traditional methods BM and SGBM also produce poor disparity maps that are too cluttered to be discriminated. This is because the unsupervised methods (BM, SGBM and PASMNet) rely entirely on stereo matching, which is very difficult in real NIR images, as described in Sec.\ref{re:depth}. Dual-Net \cite{wudual} also achieves acceptable performance in lower resolution. IFAST produces reasonable and discriminative depth maps for most of the samples. }

\begin{figure}[t]
	\begin{center}
		\includegraphics[width=\linewidth]{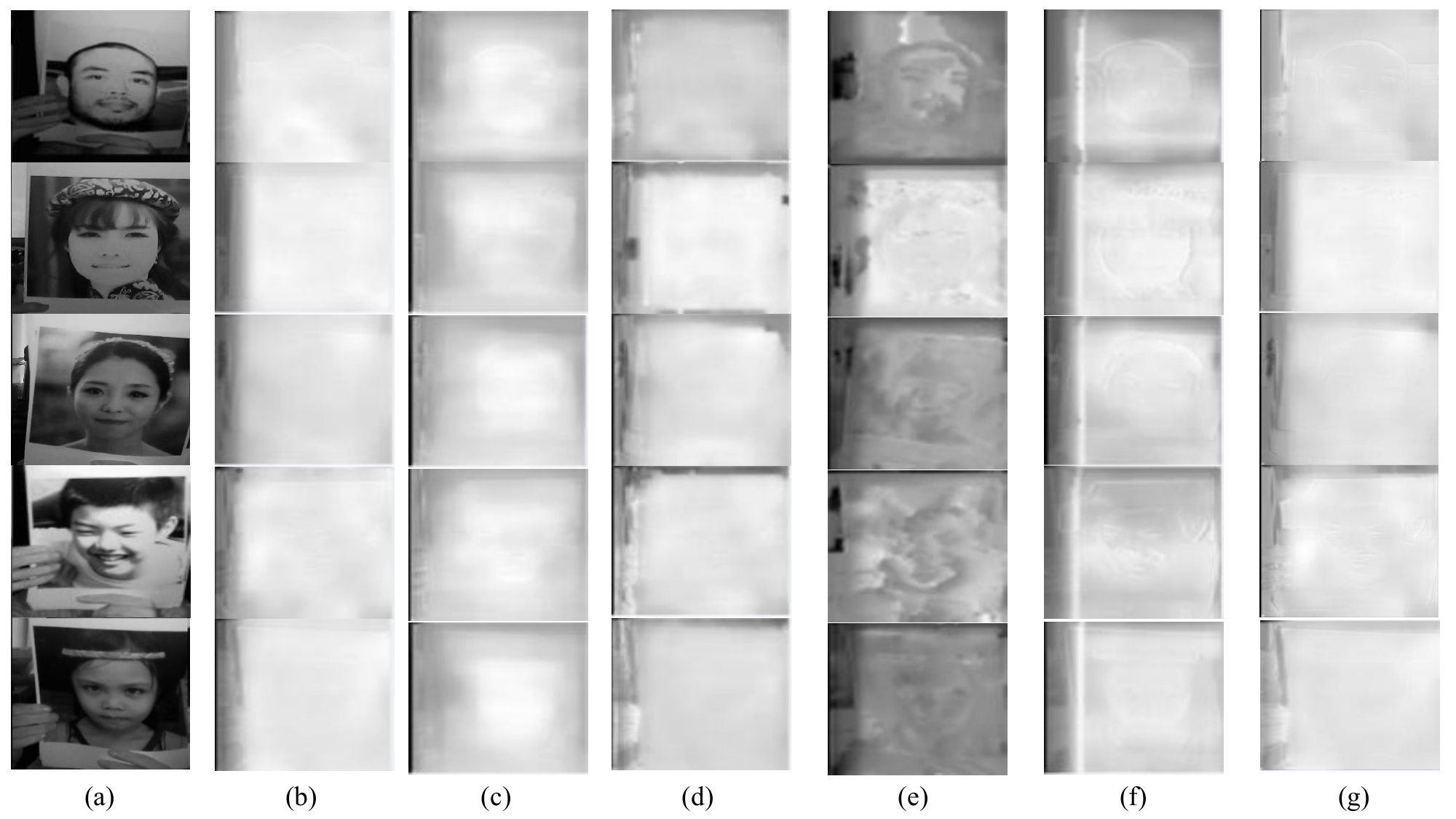}
	\end{center}
        % \vspace{-14pt}
	\caption{The disparity maps predicted on different ablation studies for plane attacks. (a) The left image. (b) Loss 1. (c) Loss 2. (d) Loss 3. (e) Up/Down sampling. (f) DMA block. (g) IFAST.}
	\label{absattack}
    % \vspace{-10pt}
\end{figure}
\begin{figure}[t]
	\begin{center}
		\includegraphics[width=\linewidth]{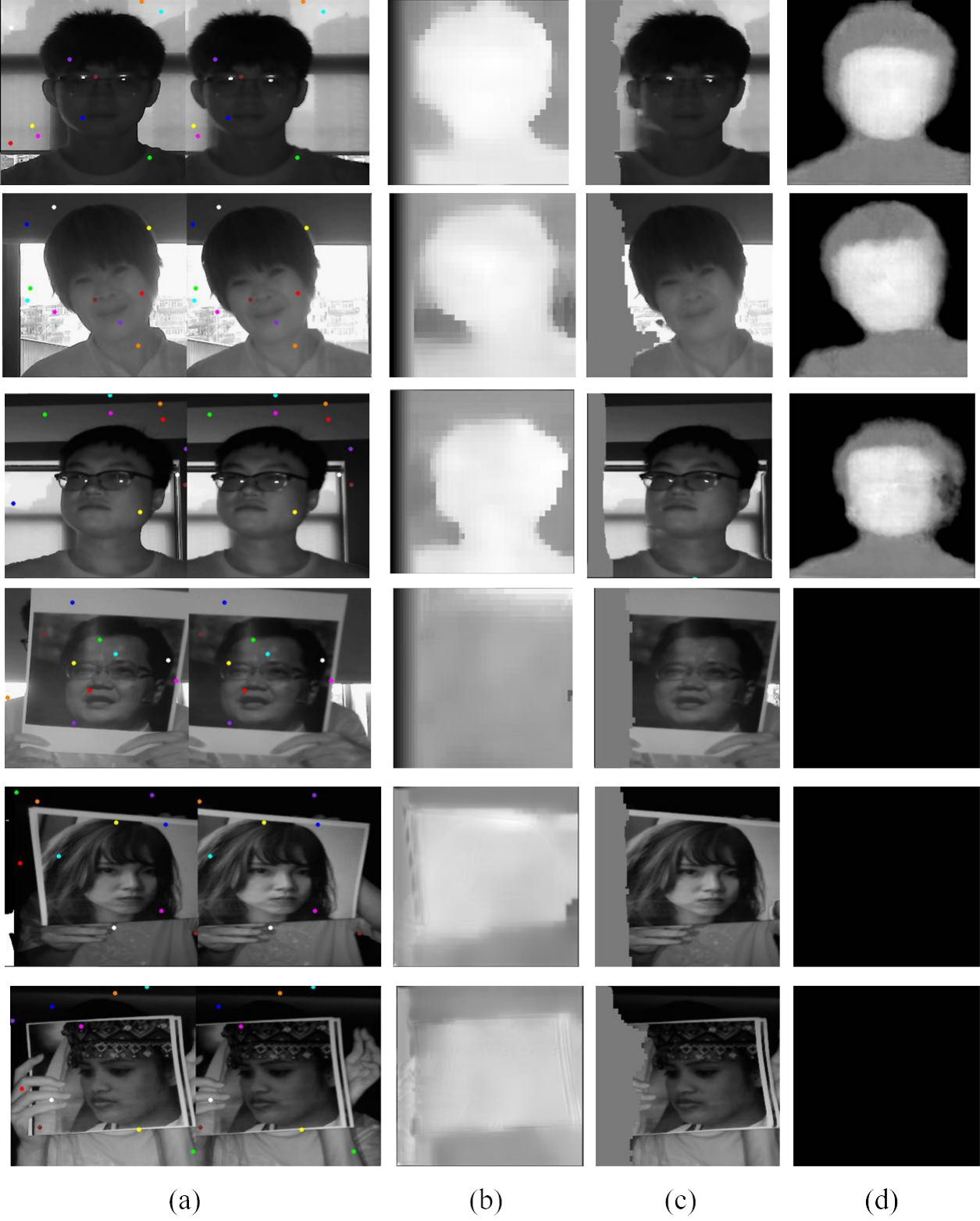}
	\end{center}
    % \vspace{-18pt}
	\caption{(a) The binocular NIR image. Note that the same colored points are the corresponding maximum attention weight points between the left and right images.  (b) The raw disparity map. (c) Reconstructed left image. (d) The confidence map on real face.} 
	\label{attn}
    % \vspace{-15pt}
\end{figure}
% \vspace{-6pt}

\subsubsection{Discussion on Attention and Stereo Matching} 
More stereo matching results based on attention weights are shown in Fig. \ref{attn}, from which it clear that the attention weights obtained by the DMA block are indeed able to achieve stereo matching between the left and right images via finding the maximum attention weights, and these amazing results are achieved even under weak supervision without manual labels. Furthermore, the reconstructed left image shown demonstrates the accuracy of the disparity map. These results indicate the interpretability of IFAST in terms of stereo vision, rather than a black-box model that simply outputs depth maps.

\subsubsection{Disparity Estimation Transformer} 
\begin{figure}[t]
	\begin{center}
		\includegraphics[width=\linewidth]{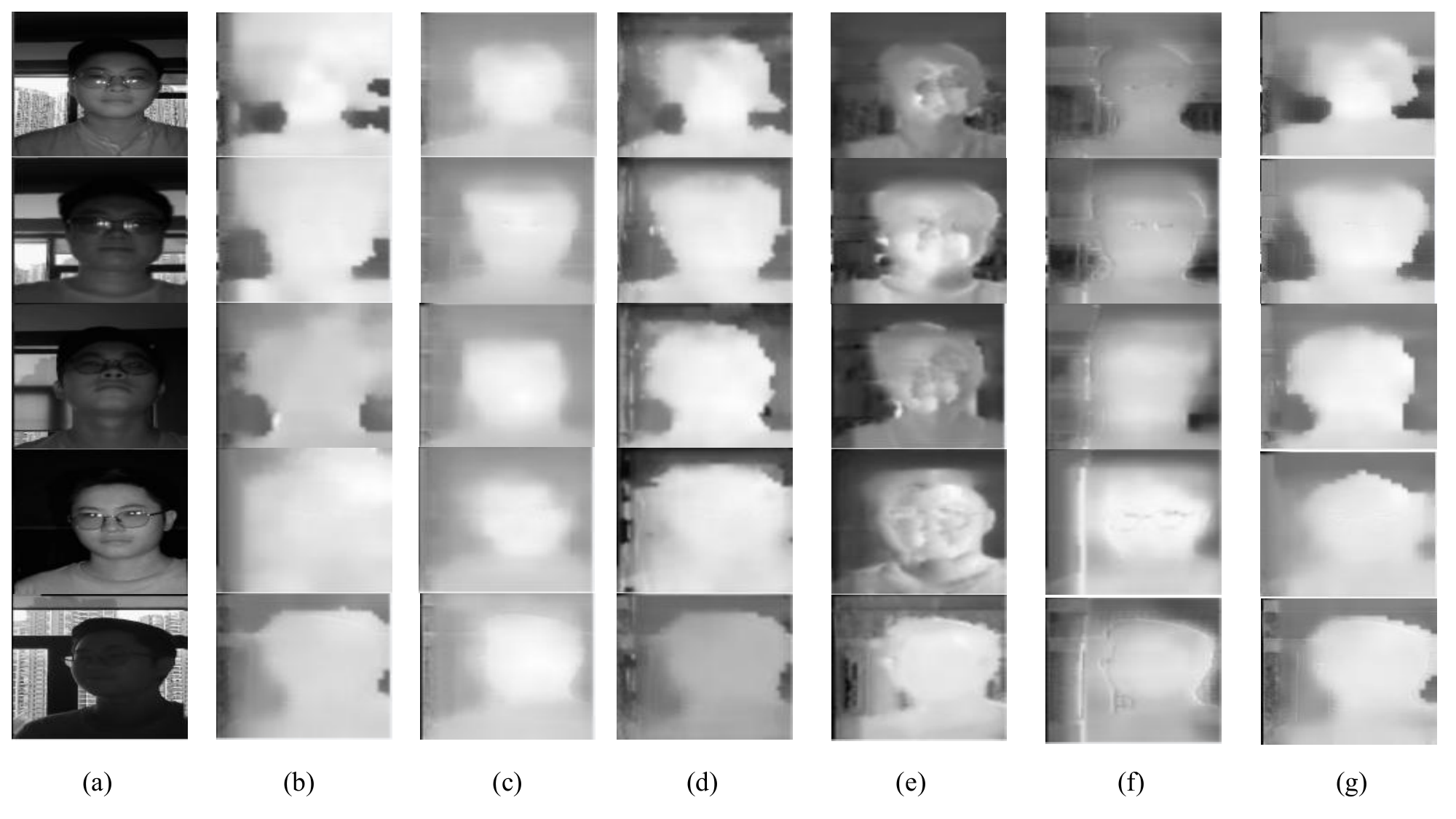}
	\end{center}
        % \vspace{-14pt}
	\caption{The disparity maps predicted in ablation studies for real faces. (a) The left image. (b) Loss 1. (c) Loss 2. (d) Loss 3. (e) Up/Down sampling. (f) DMA block. (g) IFAST.} 
	\label{absreal}
    % \vspace{-10pt}
\end{figure}

To explore the performance of our proposed DMA block and disparity estimation Transformer, some components are replaced with baseline operations. For the DMA blocks, they are replaced by simply computing self-attention and cross-attention alternately. For the up/down sampling between DMA blocks, they are removed directly. In Table \ref{abs}, it can be observed that after removing the DMA blocks or up/down sampling respectively, there is a certain decrease in the performance of IFAST. In the absence of the DMA block, the information within the left image and the mutual information between the left and right images could be easily mixed and interfered with each other in the early stages of training, as there is no pixel-wise supervision to guide the flow of information. In addition, this is equivalent to using the same window for each layer. This fixed window size would have a great impact on the matching performance. As shown in Fig.~\ref{absreal} and Fig.~\ref{absattack}, the DMA block and up/down sampling has some effect on visual depth estimation. In Fig.~\ref{absreal} (e), since the window size is fixed to $4\times 4$ during matching, the estimated depth maps for real faces and plane attacks are sharp and the visual effect is relatively bad. After removing the DMA block, it can be seen from Fig. \ref{absreal} (f) that the estimated depth maps for both real faces and plane attacks are too smooth due to too much mutual interference between the feature information of the left and right images.
\subsubsection{Confidence Map Generator}
To explore the performance of our confidence map generator as the classification part, it is replaced with other networks. First, we design another dual-branch non-gated network (called non-gated net for short), whose structure is highly similar to the proposed dual-branch gated network (called gated net for short). In non-gated net, there is no dot product operation between the two branches, nor is there a feature map concatenation. Therefore, no information is exchanged between these two branches until their classification scores are linearly combined to complete the classification. As shown in Table \ref{abs}, this non-gated net can still achieve good performance, but its blunt use of depth information and texture information makes it less effective than the gated net. In addition, the discriminability of the disparity map for classification is verified. We replace our gated net with ResNet18 \cite{he2016residual} and employ only the disparity map for classification. In Table \ref{abs}, the performance of using only the disparity map for classification is already very good, demonstrating the discriminability of the disparity map for face anti-spoofing. The gated net integrates the information of texture and disparity map, so it takes into account more discriminative features than using only disparity maps. 

\begin{figure}[t]
	%\hspace{0.5cm}
	\centering
	\includegraphics[width=0.8\linewidth]{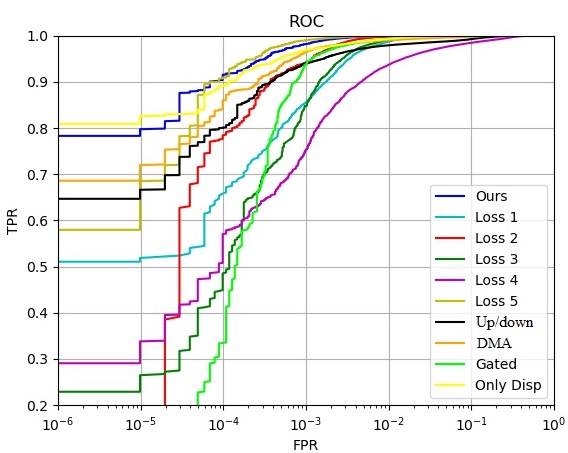}
        % \vspace{-10pt}
	\caption{The ROC curves of ablation studies for our methods, and "Loss 1" to "Loss 5" refer to remove each loss in Table \ref{abs} in turn. "Up/down", "DMA" and "Gated" refer to remove up/down sampling, DMA block, gated in CMG. "Only Disp" means that only the disparity map is fed in CMG.}
	\label{rocab}
    % \vspace{-15pt}
\end{figure}

\subsubsection{Weakly Supervised by Dual-teacher Distillation} 

Ablation experiments are performed on our weakly supervised losses. In this evaluation, 5 models with different losses are trained and measured. As shown in Table~\ref{abs} and Fig.~\ref{rocab}, all loss terms are important for the training of IFAST. Of these, the weighted relative disparity loss, reconstruction loss, and disparity smooth loss are crucial for training the depth estimation part. Without them, the FAS performance decreases significantly, especially in terms of TPR when FPR is less than 0.05\%. Focal confidence map loss and confidence map triplet loss are mainly used to train the classification part. ACC and various TPR metrics decreases by more than 10\% without them. This comparison verifies the effectiveness of the dual-teacher distillation module. It is intuitive to observe the estimated depth maps in the ablation studies for each loss. Loss 1 to loss 3 in Fig. \ref{absreal} (b), (c), and (d) refer to weighted relative disparity loss, reconstruction loss, and disparity smooth loss, respectively. In Fig. \ref{absreal} (b), the estimated depth maps are overly smooth between the foreground and background of the real face without weighted relative disparity loss, making it difficult to distinguish. In Fig. \ref{absreal} (c), although the estimated depth maps still satisfy the relative depth relation without using reconstruction loss, the depth values are often inappropriate, resulting in the inability to satisfy the pixel correspondence between the left and right images. In Fig. \ref{absreal} (d), without disparity smooth loss, the estimated depth maps are similar to the depth map estimated by IFAST, and using this loss can slightly improve the details of the depth map.

\section{Conclusion}
This study focuses on single-shot t binocular disparity-based face anti-spoofing, and Interpretable FAS Transformer (IFAST) and a large-scale binocular near-infrared image dataset (i.e., BNI-FAS) are proposed. The results show that the proposed method is effective, interpretable, and does not require manual labeling. However, our method still has many deficiencies, such as poor depth estimation performance for weakly textured regions and darkly illuminated regions, which may serve as an interesting topic that we aim to address in the future.

\bibliographystyle{IEEEtran}
\bibliography{tifs.bib}
\vfill

\clearpage
\appendix

\subsection{Details of BNI-FAS}
In session 1, there are four types of shooting conditions for real faces: illumination, distance, identity (ID), and the presence of a white box. There are three types of illumination conditions: full-light, weak-light, and night-light. The shooting distances range from 0.5 to 1.5 meters from the camera. Since the correlation between depth estimation and ID is not significant, only 6 people are selected. The presence of a white box refers to whether the person being shot is holding a white box, which examines whether the depth estimation is related to the edge of a rectangle. The shooting conditions of the plane attacks are divided into four types: illumination, distance, paper material, and the presence of a white box. In the pre-experiment, the anti-spoofing performance has a strong relationship with paper material, so three kinds of printing paper are selected, namely RGB-A4 paper, Black-A4 paper, and NIR-A4 paper. The white box refers to the white edge of the paper to investigate whether the method is affected by the white edge of the paper.

In session 2, we focus on increasing the intensity of illumination, so there are 5 types of illumination conditions: full-light, half-light, normal-light, low-light, and night-light. The shooting distances range from 0.5 meters to 2 meters. We choose 8 materials of the printed papers, namely RGB-A4 paper, RGB-IceWhite paper, RGB-Coated paper, NIR-A4 paper, NIR-Coated paper, NIR-IceWhite paper, Black-A4 paper, and Black NIR-A4 paper. The diversity of materials is a great advantage of our dataset.

\begin{figure*}[t]
	%\hspace{0.5cm}
	\centering
    \begin{minipage}[t]{0.24\linewidth}
	\subfigure[]{\includegraphics[width=\linewidth]{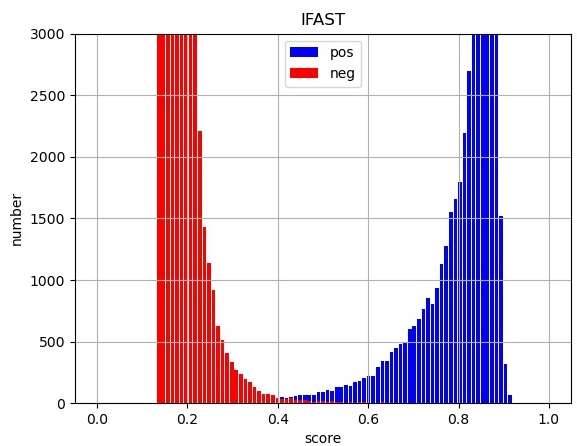}}
    \end{minipage}
    \begin{minipage}[t]{0.24\linewidth}
	\subfigure[]{\includegraphics[width=\linewidth]{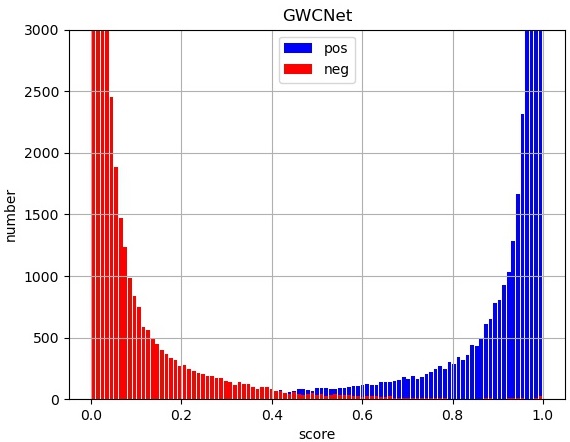}}
    \end{minipage}
    \begin{minipage}[t]{0.24\linewidth}
	\subfigure[]{\includegraphics[width=\linewidth]{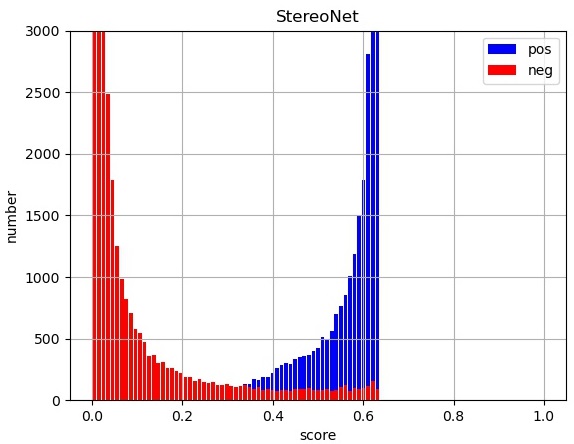}}
    \end{minipage}
	\begin{minipage}[t]{0.24\linewidth}
    \subfigure[]{\includegraphics[width=\linewidth]{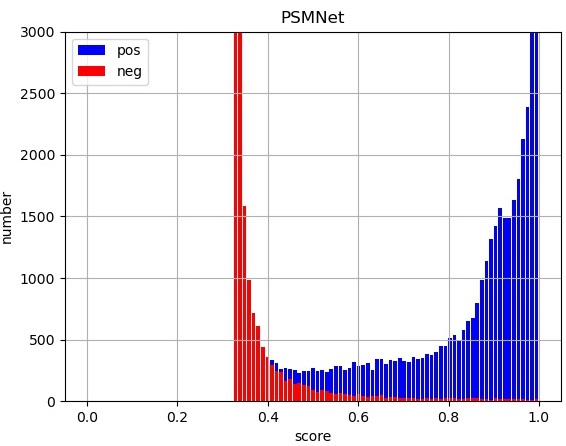}}
    \end{minipage}\\
    \begin{minipage}[t]{0.24\linewidth}
	\subfigure[]{\includegraphics[width=\linewidth]{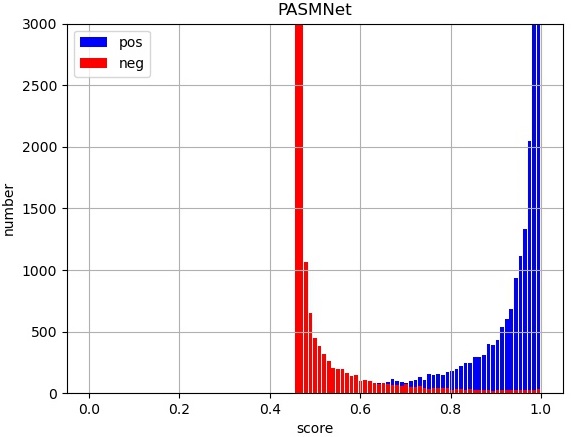}} 
    \end{minipage}
    \begin{minipage}[t]{0.24\linewidth}
	\subfigure[]{\includegraphics[width=\linewidth]{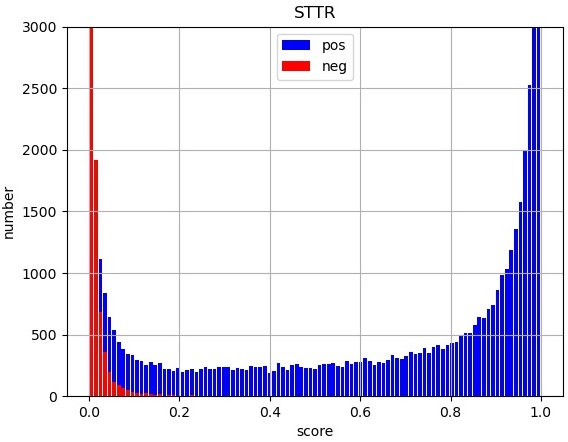}}
    \end{minipage}
    \begin{minipage}[t]{0.24\linewidth}
	\subfigure[]{\includegraphics[width=\linewidth]{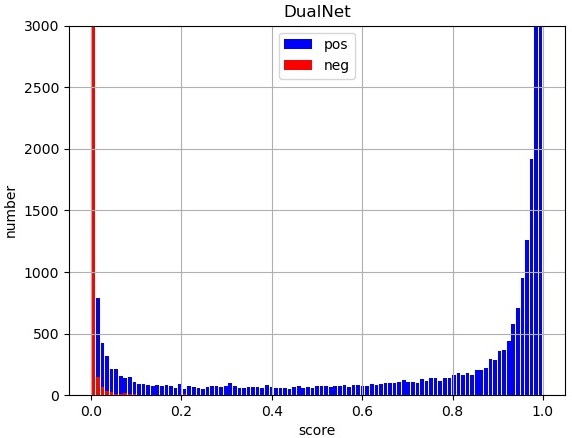}}
    \end{minipage}
    \begin{minipage}[t]{0.24\linewidth}
	\subfigure[]{\includegraphics[width=\linewidth]{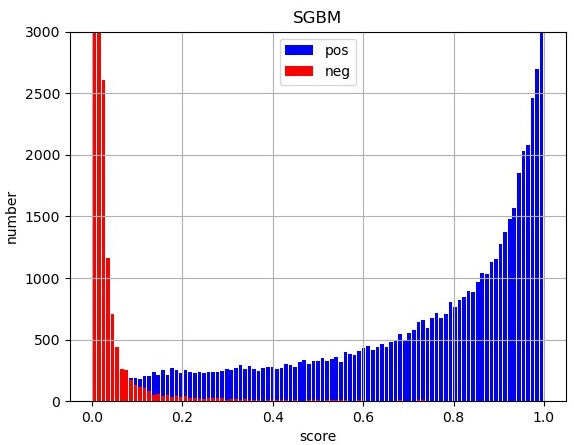}}
    \end{minipage}
	\caption{The histogram of the scores (the probability of being positive samples) of all samples in the test set by different methods, where blue represents real faces (positive samples) in the test set, and red represents plane attacks (negative samples) in the test set. (a) IFAST. (b) GWCNet (c) StereoNet. (d) PSMNet. (e) PASMNet. (f) STTR. (g) Dual-Net. (h) SGBM.} 
	\label{scoredepthfig}
\end{figure*}

\subsection{More Details of Experiment Settings}
\subsubsection{Comparison Experiments of Different Depth Estimation Methods}\label{exper1}
For the training set, we select about 40,000 images from 2 IDs as positive samples, and select about 40,000 plane attack images under various conditions as negative samples. For the test set, we select about 56,000 images from the other 3 IDs as positive samples, and select about 100,000 plane attack images under various conditions as negative samples.

\subsubsection{Model Generalization Experiments on Different Test Settings}\label{exper2}
In the experiment to compare different test settings, we first fix our training set as follows: all images of ID M in session 2 are taken as positive samples, and all attack images of Black-A4 material in session 2 are taken as negative samples. We then selectively change the test set to evaluate performance under specific conditions. In the experiment to compare paper attacks with different materials, the training set is as follows: the daytime images of ID M in session 2 are used as positive samples, and the plane attacks of all Black-A4 materials in session 2 are used as negative samples. Then we use all the images of ID W as the positive samples of the test set and selectively change the negative samples in the test set to evaluate the performance under different materials.

\begin{figure}[t]
	%\hspace{0.5cm}
	\centering
    \begin{minipage}[t]{0.48\linewidth}
	\subfigure[]{\includegraphics[width=\linewidth]{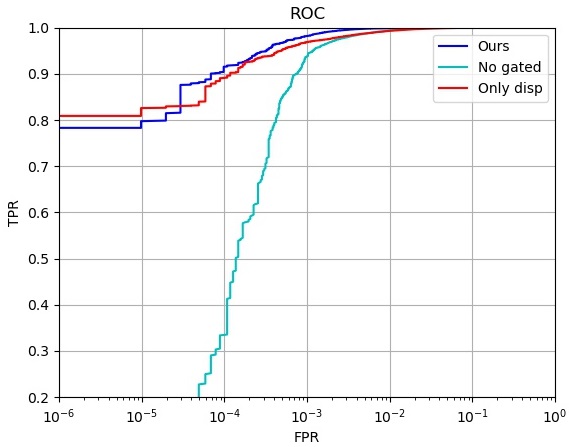}}
    \end{minipage}
    \begin{minipage}[t]{0.48\linewidth}
	\subfigure[]{\includegraphics[width=\linewidth]{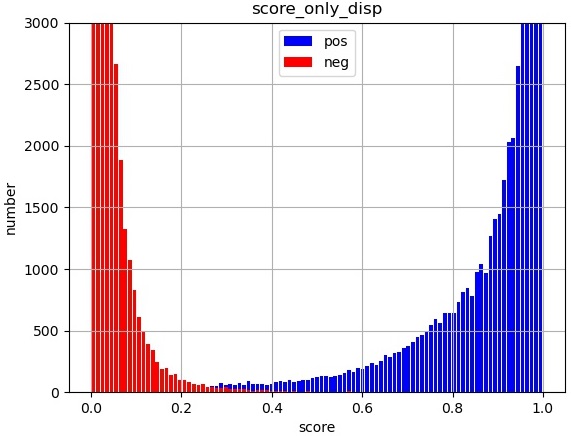}}
    \end{minipage}
	\caption{(a) The ROC curves of ablation studies for our CMG. "NO gated” refers to the non-gated branch net for replacing CMG. ”Only disp” means replacing the CMG with ResNet18 and using only disparity for classification. (b) The score histogram of "Only disp".} 
	\label{com_cmg}
\end{figure}

\subsection{More Score Histogram in Experiments}
\subsubsection{Comparison of Different Depth Estimation Methods}
For further analysis, although all methods achieve satisfactory FAS performance by using our confidence map generator. However, as can be seen in Fig.~\ref{scoredepthfig}, different methods still have significant differences in the score distribution of the test samples, where we define the score here as the probability of predicting that a sample is a real face. An ideal score distribution should be: positive samples (blue in Fig.~\ref{scoredepthfig}) are mostly concentrated on the right with high scores, while negative samples (red) are mostly concentrated on the left with low scores, and there should be a clear boundary between positive and negative samples. Among all the compared methods, IFAST, GWCNet, STTR, Dual-Net, and SGBM have relatively reasonable score distributions. However, in the test results of STTR and SGBM, there are many positive samples distributed in low-scoring areas. The score distributions of StereoNet, PSMNet, and PASMNet are obviously skewed in one direction. We believe that if the overall effect of depth estimation is biased toward smooth or sharpness, it may affect the score distribution for classification, i.e., it tends to give high or low scores to all samples.

\begin{figure}[t]
	%\hspace{0.5cm}
	\centering
	\subfigure[]{\includegraphics[width=0.49\linewidth]{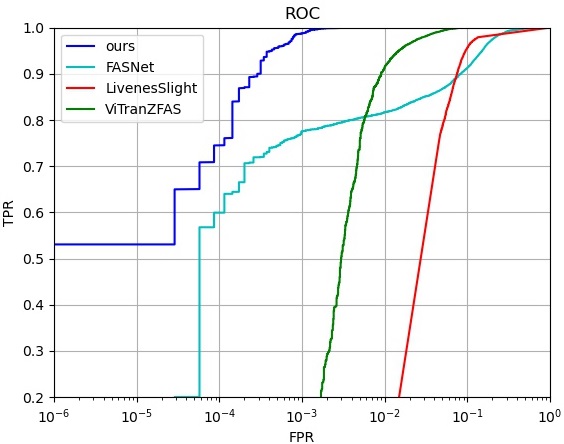}}
	\subfigure[]{\includegraphics[width=0.49\linewidth]{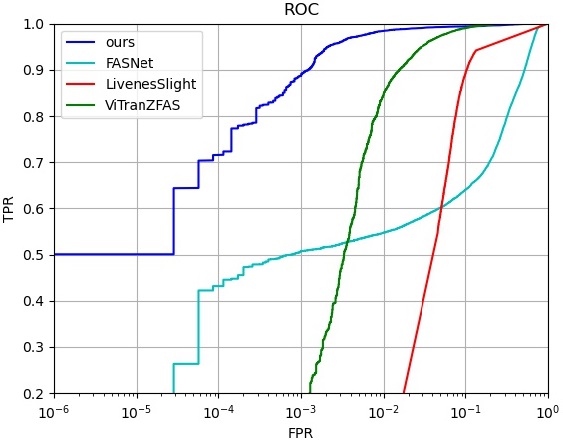}}
	\caption{The classification ROC curves of different methods. (a)Test set (positive samples: Night-Light in Session 1, negative samples: ColorA4 in Session 2). (b)Test set (positive samples: Weak-Light in Session 2, negative samples: ColorA4 in Session 2) } 
	\label{rocfig2}
\end{figure}

\subsubsection{Comparison of Model Generalization on Different Test
Settings}
\begin{figure*}[t]
	%\hspace{0.5cm}
	\centering
 \begin{minipage}[t]{0.24\linewidth}
	\subfigure[]{\includegraphics[width=\linewidth]{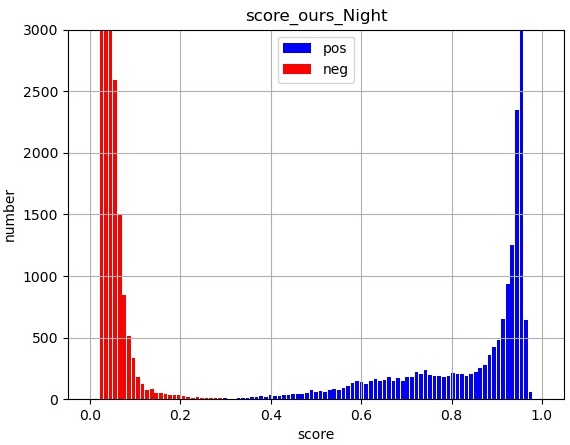}}
    \end{minipage}
    \begin{minipage}[t]{0.24\linewidth}
	\subfigure[]{\includegraphics[width=\linewidth]{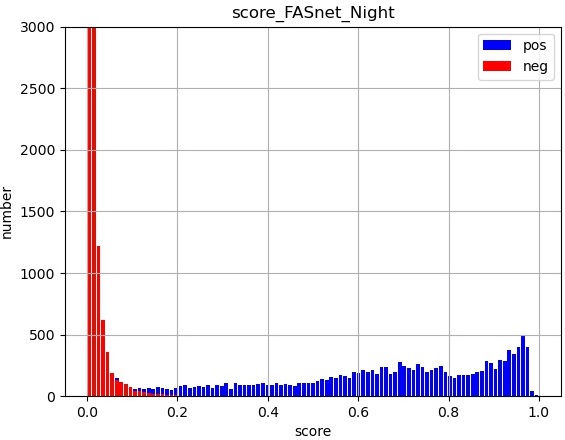}}
    \end{minipage}
    \begin{minipage}[t]{0.24\linewidth}
	\subfigure[]{\includegraphics[width=\linewidth]{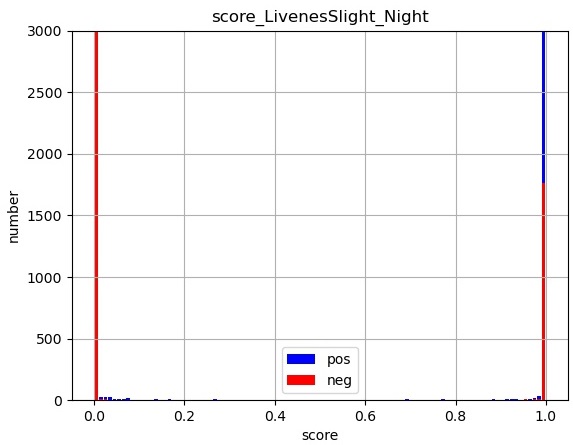}}
    \end{minipage}
	\begin{minipage}[t]{0.24\linewidth}
    \subfigure[]{\includegraphics[width=\linewidth]{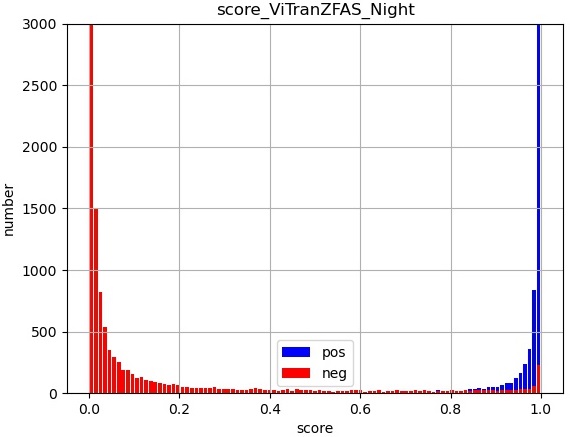}}
    \end{minipage}
	\caption{The score histogram of all samples in the test set (Positive samples: Night-Light in session 2, Negative samples: ColorA4 in session 2) predicted as real faces by different methods. (a) IFAST. (b) FASNet (c) LivenesSlight. (d) VitranZFAS.} 
	\label{scorenightfig}
\end{figure*}

\begin{figure*}[t]
	%\hspace{0.5cm}
	\centering
    \begin{minipage}[t]{0.24\linewidth}
	\subfigure[]{\includegraphics[width=\linewidth]{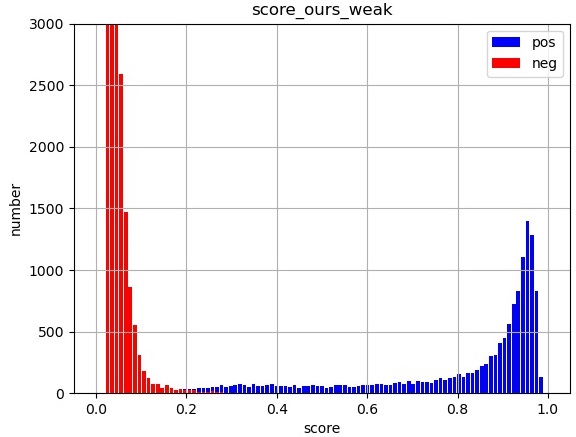}}
    \end{minipage}
    \begin{minipage}[t]{0.24\linewidth}
	\subfigure[]{\includegraphics[width=\linewidth]{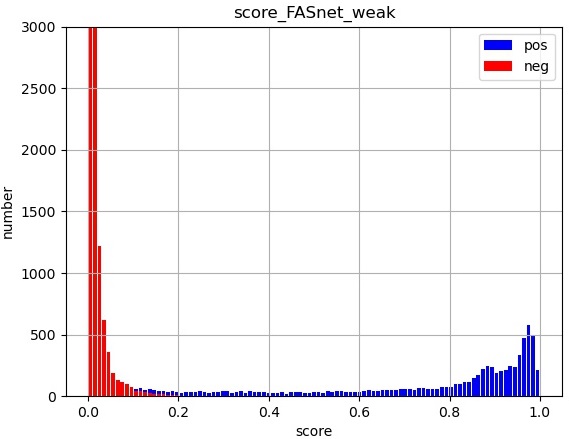}}
    \end{minipage}
    \begin{minipage}[t]{0.24\linewidth}
	\subfigure[]{\includegraphics[width=\linewidth]{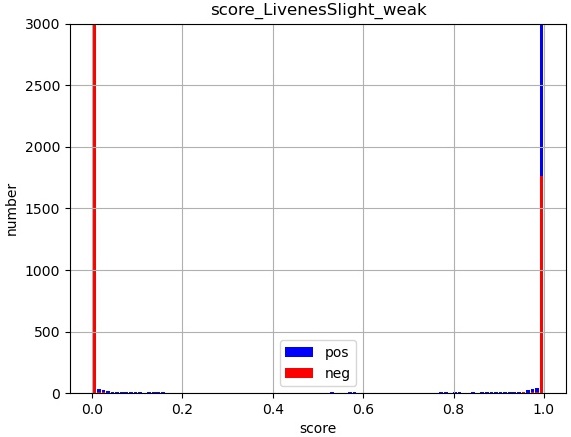}}
    \end{minipage}
	\begin{minipage}[t]{0.24\linewidth}
    \subfigure[]{\includegraphics[width=\linewidth]{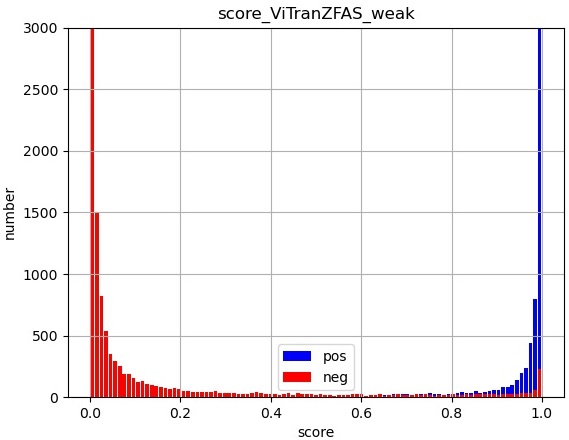}}
    \end{minipage}
	\caption{The score histogram of all samples in the test set (Positive samples: Weak-Light in session 1, negative samples: ColorA4 in session 2) predicted as real faces by different methods. (a) IFAST. (b) FASNet (c) LivenesSlight. (d) VitranZFAS.} 
	\label{scoreweakfig}
\end{figure*}

For further analysis, we focus on the effect of night-light on IFAST and compare monocular methods. As shown in Fig. \ref{scorenightfig}, when we set the test set as follows: night-light images in session 2 as positive samples and RGB-A4 images in session 2 as negative samples, there is a drop in performance for all methods. The evaluation scores of IFAST are relatively high for most of the night-light positive samples. In contrast, the performance collapse of the monocular methods varies. FASNet scores low for most of the night-light positive samples (as in Fig. \ref{scorenightfig} (b) that the number of positive samples on the right is significantly small), while LivenesSlight scores high for many negative samples (as in Fig. \ref{scorenightfig} (c) that a large number of negative samples also receive extremely high scores). The score distribution of VitranZFAS is relatively good, but there are also many negative samples that receive high scores. The score distributions of weak-light images are similar to those of the night-light images for all methods in Fig. \ref{scoreweakfig}. The ROC curves under these test settings are given in Fig. \ref{rocfig2}. If the score distribution is unreasonable (positive samples get low scores or negative samples get high scores), it will have a huge impact on the ROC curve. In particular, if negative samples get high scores, the TPR (for a given FPR) will drop very quickly.
\begin{table}[t]
\centering
\scalebox{0.6}{
\begin{tabular}{cccccccc}
\hline
Image   & StereoNet & PSMNet & GwcNet & STTR & PASMNet & Dual-Net & IFAST \\
\hline
1       & 5.56 & 6.28 & 7.23 & 6.55 & 6.37 & 6.67 & \textbf{7.68} \\
2       & 6.12 & 7.12 & 7.12 & 6.91 & 5.91 & 7.57 & \textbf{8.44} \\
3       & 5.86 & 6.69 & 7.01 & 5.71 & 4.90 & 7.24 & \textbf{8.00} \\
4       & 5.13 & 6.56 & 5.68 & 5.61 & 6.54 & 7.52 & \textbf{7.81} \\
5       & 5.80 & 6.96 & 6.23 & 5.78 & 6.15 & 7.10 & \textbf{7.84} \\
6       & 4.78 & 6.44 & 6.73 & 5.53 & 5.80 & 6.89 & \textbf{8.01} \\
7       & 5.65 & 5.95 & 6.24 & 5.40 & 6.08 & 6.11 & \textbf{7.76} \\
8       & 4.42 & 5.33 & 6.06 & 4.28 & 4.79 & 5.49 & \textbf{6.58} \\
9       & 5.78 & 6.75 & 6.02 & 5.52 & 5.58 & 7.00 & \textbf{7.36} \\
10      & 6.44 & 7.32 & 6.10 & 6.34 & 5.12 & 5.04 & \textbf{8.12} \\
\hline
average &5.55 & 6.54 & 6.44 & 5.76 & 5.72 & 6.66 &\textbf{7.76}\\
\hline
\end{tabular}
}
\caption{The scores of 10 estimated depth images for user study.}\label{tab:user-study}
\end{table}
\subsection{Ablation Studies}
\begin{figure*}[!ht]
	%\hspace{0.5cm}
	\centering
    \begin{minipage}[t]{0.24\linewidth}
	\subfigure[]{\includegraphics[width=\linewidth]{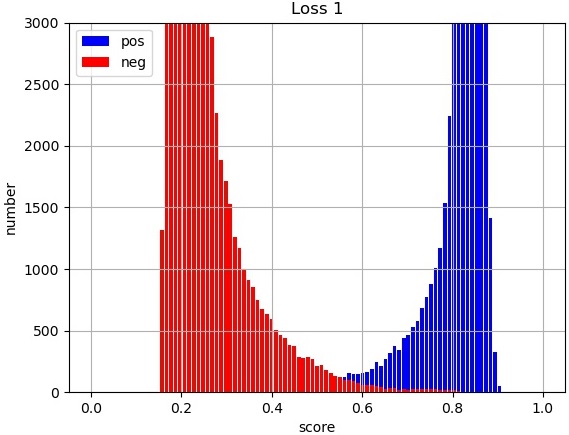}}
    \end{minipage}
    \begin{minipage}[t]{0.24\linewidth}
	\subfigure[]{\includegraphics[width=\linewidth]{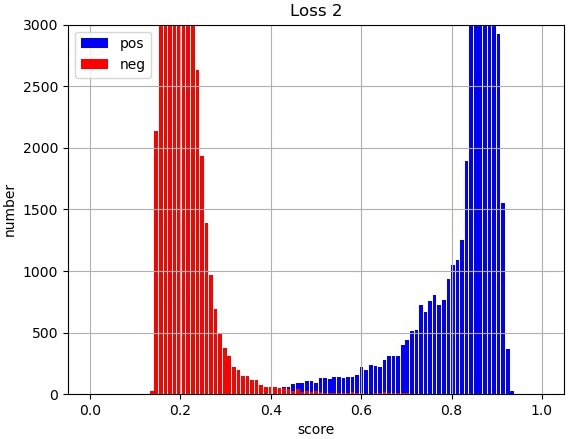}}
    \end{minipage}
    \begin{minipage}[t]{0.24\linewidth}
	\subfigure[]{\includegraphics[width=\linewidth]{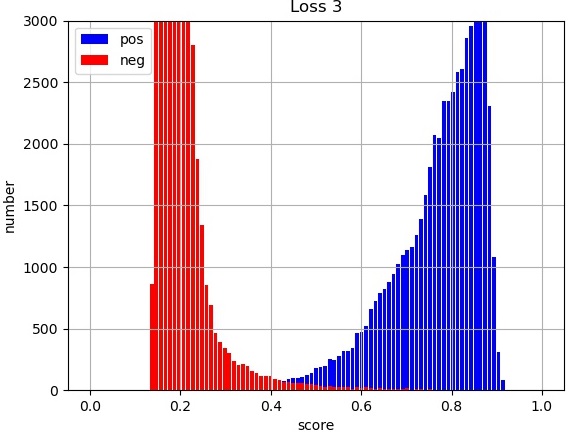}}
    \end{minipage}
	\begin{minipage}[t]{0.24\linewidth}
    \subfigure[]{\includegraphics[width=\linewidth]{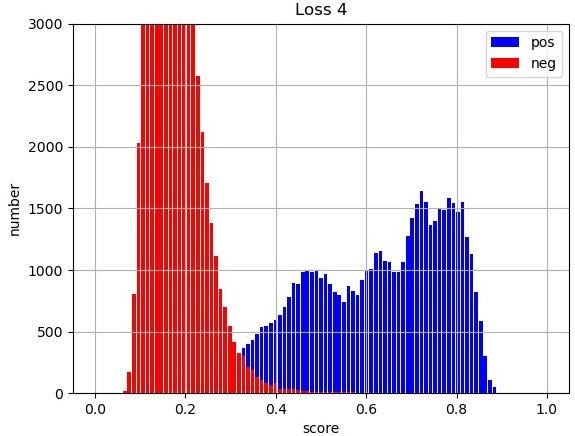}}
    \end{minipage}\\
    \begin{minipage}[t]{0.24\linewidth}
	\subfigure[]{\includegraphics[width=\linewidth]{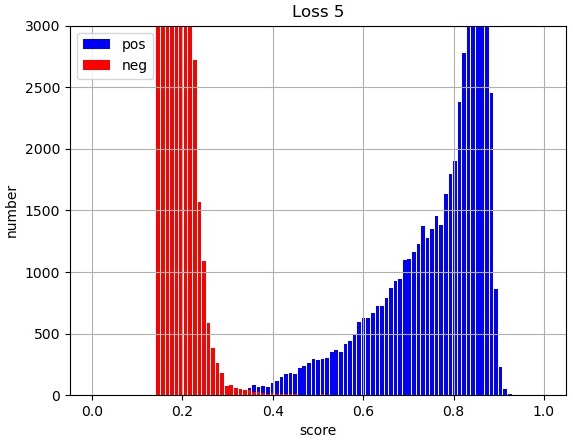}}
    \end{minipage}
    \begin{minipage}[t]{0.24\linewidth}
	\subfigure[]{\includegraphics[width=\linewidth]{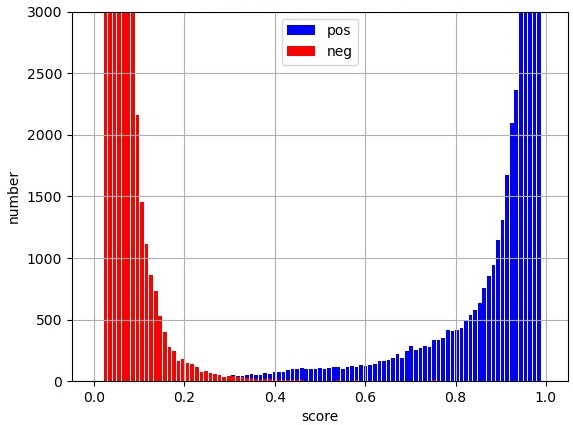}}
    \end{minipage}
    \begin{minipage}[t]{0.24\linewidth}
	\subfigure[]{\includegraphics[width=\linewidth]{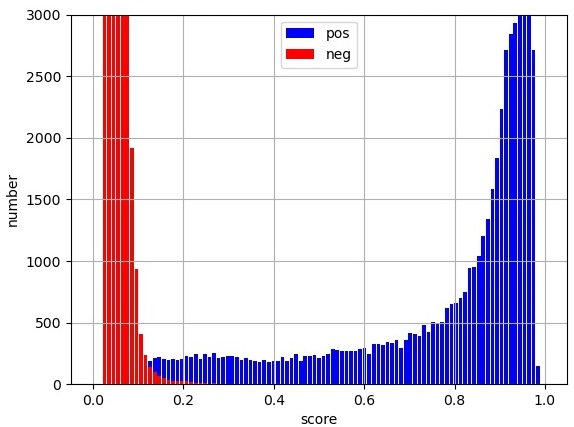}}
    \end{minipage}
	\begin{minipage}[t]{0.24\linewidth}
    \subfigure[]{\includegraphics[width=\linewidth]{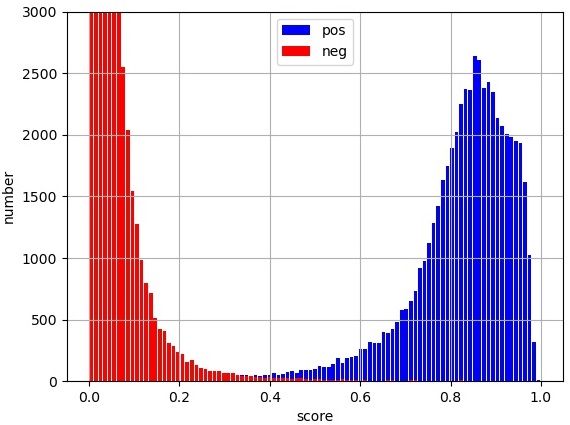}}
    \end{minipage}
	\caption{The histogram of the scores of all samples in the test set predicted as real faces on different ablation studies, where blue represents real faces in the test set (positive samples), and red represents plane attacks (negative samples). (a) Loss 1. (b) Loss 2. (c) Loss 3. (d) Loss 4. (e) Loss 5. (f) DMA block. (g) Up/down sampling. (h) Gated in CMG.} 
	\label{scoreabs}
\end{figure*}

In terms of details, as can be seen from Fig. \ref{scoreabs} (a), without relative disparity loss makes some negative samples get high scores. In Fig. \ref{scoreabs} (b), (c), and (e), without reconstruction loss, disparity smooth loss, or confidence map triplet loss make a number of positive samples get lower scores. Surprisingly, without focal confidence map loss, the performance collapses significantly, and the scores of a large number of positive samples become very low in Fig. \ref{scoreabs} (d). It can be seen that the focal confidence map loss is a very important way of knowledge transfer in dual-teacher distillation.


\section{User Study}
We further conduct a user study with 30 participants to evaluate the human perception of our method on 10 estimated depth maps. The participants are asked to score the visual quality of the estimated depth maps from 1 (worst) to 9 (best). Table \ref{tab:user-study} reports the average rating scores of different methods, among which our method receives the highest ratings.

\end{document}